\documentclass[10pt,journal,compsoc]{IEEEtran}

\ifCLASSOPTIONcompsoc
  \usepackage[nocompress]{cite}
\else
  \usepackage{cite}
\fi

\usepackage{wrapfig}
\usepackage{graphicx}
\usepackage[table]{xcolor}
\definecolor{LightCyan}{rgb}{0.88,1,1}
\definecolor{LightYellow}{rgb}{1,1,0.7}
\usepackage{verbatim}
\usepackage{amsmath}
\usepackage{hyperref}
\usepackage[percent]{overpic}
\usepackage{pifont}
\usepackage{bbding}
\usepackage{pifont}
\usepackage{wasysym}
\usepackage{amssymb}
\usepackage{xcolor}
\usepackage{bm}

\newcommand{\matteo}[1]{ \color{black} #1 \color{black}}

\definecolor{LightYellow}{rgb}{1,1,0.7}

\def\eg{\emph{e.g. }}

\def\ie{\emph{i.e. }}
\def\etal{\emph{et al. }}
\begin{document}
\title{On the Synergies between Machine Learning and Binocular Stereo for Depth Estimation from Images: a Survey}

\author{Matteo~Poggi,~\IEEEmembership{Member,~IEEE,}
        Fabio~Tosi,~\IEEEmembership{Student~Member,~IEEE,}
        Konstantinos~Batsos,~\IEEEmembership{Student~Member,~IEEE,}\\
        Philippos~Mordohai,~\IEEEmembership{Member,~IEEE,}
        and~Stefano~Mattoccia,~\IEEEmembership{Member,~IEEE}
\IEEEcompsocitemizethanks{\IEEEcompsocthanksitem M. Poggi, F. Tosi and S. Mattoccia are with the Department
of Computer Science and Engineering, University of Bologna, Italy,
IT.\protect\\
$\lbrace$m.poggi,fabio.tosi5,stefano.mattoccia$\rbrace$ @unibo.it
}

\IEEEcompsocitemizethanks{\IEEEcompsocthanksitem K. Batsos is with Argo AI. \protect\\ kbatsos@stevens.edu
}

\IEEEcompsocitemizethanks{\IEEEcompsocthanksitem P. Mordohai is with the Department
of Computer Science, Stevens Institute of Technology, New Jersey,
USA.\protect\\
philippos.mordohai@stevens.edu
}
}


\IEEEtitleabstractindextext{
\begin{abstract}
Stereo matching is one of the longest-standing problems in computer vision with close to 40 years of studies and research.
Throughout the years the paradigm has shifted from local, pixel-level decision to various forms of discrete and continuous optimization to data-driven, learning-based methods. 
Recently, the rise of machine learning and the rapid proliferation of deep learning enhanced stereo matching with new exciting trends and applications unthinkable until a few years ago.
Interestingly, the relationship between these two worlds is two-way. While machine, and especially deep, learning advanced the state-of-the-art in stereo matching, stereo itself enabled new ground-breaking methodologies such as self-supervised monocular depth estimation based on deep networks.
In this paper, we review recent research in the field of learning-based depth estimation from single and binocular images highlighting the synergies, the successes achieved so far and the open challenges the community is going to face in the immediate future. 
\end{abstract}
}

\maketitle

\begin{figure*}
    \centering
    \renewcommand{\tabcolsep}{1pt}
    \begin{tabular}{ccc}
        {\includegraphics[width=0.3\textwidth]{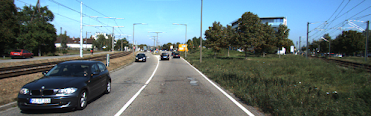}} &
        \begin{overpic}[width=0.3\textwidth]{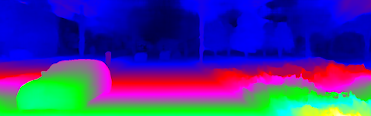}
        \put (2,25) {$\displaystyle\textcolor{white}{\textbf{a)}}$}
        \end{overpic} &
        \begin{overpic}[width=0.3\textwidth]{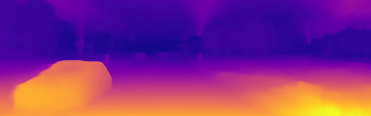}
        \put (2,25) {$\displaystyle\textcolor{white}{\textbf{b)}}$}
        \end{overpic} \\
    \end{tabular}
    \caption{\textbf{Years of progress in the field of stereo vision and machine learning} enable the estimation of depth maps of unprecedented quality from a) stereo or b) monocular images.}
    \label{fig:teaser}
\end{figure*}

\IEEEdisplaynontitleabstractindextext

\IEEEpeerreviewmaketitle

\section{Introduction}

Since the early stages of computer vision, estimating depth from images has been one of the iconic challenges for researchers. 
Obtaining dense and accurate depth maps is crucial for effectively addressing higher-level tasks such as 3D reconstruction, mapping and localization, autonomous driving, and many more. 
The focus of this paper is on stereo matching, which is classified as a passive sensing technique, and related topics. Competing technologies for depth estimation rely on active sensing which comes in several forms, including structured light projection, Time-Of-Flight (ToF) measurement, Laser Imaging Detection and Ranging (LIDAR) among others. Common to these devices is the perturbation of the environment required to sense depth. Although very accurate and precise, these sensors suffer from non-negligible weaknesses limiting their practical deployment for real applications. For instance, LIDAR sensors, which rely on one or more laser emitters scanning the environment through mechanical rotation, may suffer from misalignment, missing laser returns due to absorbing or reflective surfaces and multi-pathing. Moreover, they typically provide only sparse measurements of the observed scene, with density (and pricing) increasing with the number of laser emitters.
For structured-light devices, such as the Microsoft Kinect, the pattern projection technology \matteo{constrains} the working range to a few meters and prevents usage under direct sunlight.

Inferring depth from images acquired by a regular camera has the potential to overcome all the limitations above. Among the different techniques for this purpose, stereo matching \cite{scharstein2002taxonomy} takes as input two rectified images and \matteo{attempts to compute} the disparity of every pixel 
by matching corresponding pixels along conjugate epipolar lines, thus enabling depth estimation via triangulation.
Years of research proved the effectiveness of stereo, making it a viable alternative to expensive active sensors often deployed in practical applications.
The success and proliferation of machine learning and deep learning techniques in computer vision \cite{lecun2015deep} led to notable improvements to stereo matching, even though it was one of the areas of computer vision in which learning was adopted relatively late.
At the same time, the most recent advances in depth estimation from images have demonstrated that deep learning itself could benefit from stereo to achieve goals unimaginable just a few years ago, as in the case of self-supervised single-image depth estimation enabled through view synthesis \cite{garg2016unsupervised} or other stereo-based strategies. Thus, the synergy between these two worlds led to outstanding results, shown in \autoref{fig:teaser}. 

In this paper, we present a comprehensive review of the last years of progress in the field of depth estimation via \matteo{binocular} stereo matching and related topics. Starting from early attempts to leverage machine learning to replace single steps of the traditional stereo pipeline \cite{scharstein2002taxonomy}, we will guide the reader through five years of research, highlighting the successes achieved so far and pointing out the open challenges the community is going to face in the immediate future. This paper extends the topics covered by \textit{Learning-based depth estimation from stereo and monocular images: successes, limitations and future challenges} tutorials offered at 3DV 2018\footnote{\url{sites.google.com/view/3dv-2018-depth-from-image}} and CVPR 2019\footnote{\url{sites.google.com/view/cvpr-2019-depth-from-image}}. We argue that it is timely since the previous surveys on stereo \cite{scharstein2002taxonomy,brown03} are outdated.

\matteo{The rest of the manuscript is organized as follows: \autoref{sec:datasets} introduces the most popular datasets and benchmarks in stereo matching, \autoref{sec:pipeline} discusses early attempts to replace individual steps of conventional stereo pipelines \cite{scharstein2002taxonomy} with learning-based techniques, followed by \autoref{sec:e2e} that reviews and classifies end-to-end models for stereo matching. Then, we consider two aspects concerning respectively the conventional pipelines and the end-to-end models, that are confidence estimation, covered in \autoref{sec:confidence}, and the domain-shift problem, introduced in \autoref{sec:shift} together with techniques aimed at mitigating it. 
Then, \autoref{sec:mono} reviews single-image depth estimation frameworks supervised by means of stereo images and finally, \autoref{sec:discussion} collects take-home messages from our survey.}

\section{Datasets}
\label{sec:datasets}

In most computer vision problems, the availability of large and diverse datasets is of paramount importance for successfully developing new algorithms and for being able to measure their effectiveness. For years, researchers in stereo matching evaluated their proposals on a few dozen stereo pairs with ground truth depth maps acquired in controlled, indoor environments \cite{scharstein2002taxonomy,scharstein2003high,HirschmullerS07}. Although these datasets allowed notable progress in the design of stereo algorithms, they did not adequately highlight many of the challenges arising in real applications. Moreover, modern machine learning algorithms are data-hungry and require much more than a few dozen stereo pairs.

\begin{figure}
    \centering
    \includegraphics[width=0.4\textwidth]{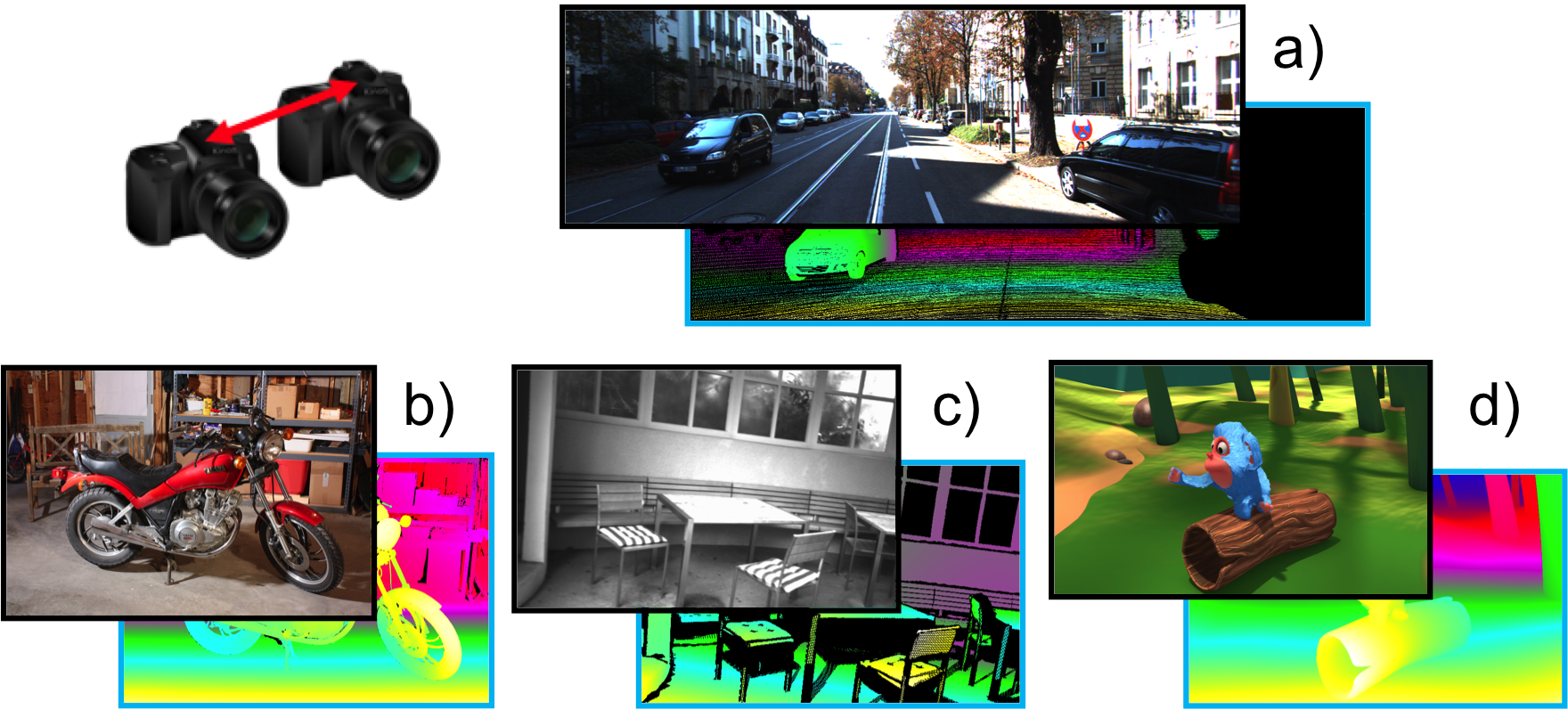}
    \caption{\textbf{Overview of the most popular stereo datasets in literature}, with examples of reference images and associated ground truth disparity. a) KITTI 2015 \cite{menze2015object}, b) Middlebury 2014 \cite{scharstein2014high}, c) ETH3D \cite{schops2017multi}, d) Freiburg SceneFlow \cite{mayer2016large}.}
    \label{fig:datasets}
\end{figure}

In 2012, the first large-scale dataset with images of outdoor, real environments was released \cite{Geiger2012CVPR} and an indoor dataset with much higher resolution \cite{scharstein2014high} appeared soon after. Later, with the advent of deep learning \cite{lecun2015deep} these datasets were followed by large, synthetic image sets which are ideal for training deep networks thanks to the negligible cost required to generate a multitude of training samples. 
In all cases, the datasets provide depth annotations obtained through different methodologies discussed later. 
The rest of this section will introduce in detail each of these datasets, summarized in \autoref{fig:datasets} where we show one reference image and the associated ground truth disparity map for each of them, respectively for a) KITTI 2015, b) Middlebury 2014, c) ETH3D and d) Freiburg SceneFlow. The first three were the foundation of the stereo aspect of the Robust Vision Challenge (ROB)\footnote{\url{robustvision.net}} in 2018.

\subsection{KITTI}

Acquired by Geiger \etal \cite{geiger2013vision}, the KITTI Vision Benchmark Suite represents the first, large-scale collection of images from a driving environment. The KITTI benchmarks have been seminal to the development of several algorithms and methods supporting autonomous driving. The data have been acquired from a car equipped with two stereo camera pairs, one grayscale and one color, a Velodyne LIDAR, GPS and inertial sensors. It consists of about 42k stereo pairs and LIDAR point clouds taken from 61 different scenes. From this extensive collection of images, appropriate benchmarks are available for key computer vision tasks such as stereo, optical flow, visual odometry, object detection and more. Two main datasets are available for stereo matching: KITTI 2012 and KITTI 2015.

\textbf{KITTI 2012 \cite{Geiger2012CVPR}. } This is the first dataset for stereo matching comprising outdoor images of static scenes and providing an online benchmark\footnote{\url{cvlibs.net/datasets/kitti/eval_stereo_flow.php?benchmark=stereo}} for evaluation. It consists of 389 grayscale stereo pairs (recently made available in color format as well), split into 194 training pairs with available ground truth and 195 test pairs with withheld ground truth.
Ground truth depth was obtained from LIDAR measurements as follows. A set of consecutive frames (5 before and 5 after) were registered using ICP, accumulated point clouds were re-projected onto the image, and finally, all ambiguous image regions such as windows and fences were manually removed. Using calibration parameters, the 3D points were projected on the images to obtain depth measurements, which were converted into disparities. This strategy yields semi-dense ground truth maps, covering about one third of the pixels in each input image.
\matteo{The error metrics on the benchmark are the percentage of pixels with a disparity error greater than 3 and the average disparity error, measured either on all pixels or non-occluded pixels only. In both cases, the lower the better.
Metrics computed in reflective regions are also available.}

\textbf{KITTI 2015 \cite{menze2015object}. }
A few years later, an improved dataset and benchmark for scene flow estimation \cite{menze2015object} was proposed. In this case, the dataset consists of 400 color stereo pairs, evenly split into training and test sets.
In contrast to the previous dataset, the stereo pairs are from dynamic scenes with objects (mostly cars) moving independently.
The same procedure used for KITTI 2012 is followed here to obtain ground truth labels, except for moving objects whose 3D points cannot be properly accumulated over time. Hence, to obtain depth annotations for cars, 3D cad models are fitted into accumulated point clouds and re-projected onto the image.
\matteo{As the primary evaluation metric, the percentage of pixels with an absolute disparity error greater than 3 and a relative error larger than 5\% (D1) is reported on the online benchmark\footnote{\url{cvlibs.net/datasets/kitti/eval_scene_flow.php?benchmark=stereo}}, the lower the better. The D1 metric is listed for 
foreground (\ie belonging to moving objects), background or all pixels. Moreover, masks to distinguish between non-occluded and all pixels are available.}

\subsection{Middlebury}

The Middlebury Stereo Vision Page provided the first benchmark that allowed authors to submit the results of their algorithms. Over the years, the Middlebury stereo datasets have provided indoor images with dense ground truth labels, obtained by manual annotation at first \cite{scharstein2002taxonomy} and by structured light sensors later \cite{scharstein2003high,HirschmullerS07,scharstein2014high}. Three main versions have been proposed between 2002 \cite{scharstein2002taxonomy} and 2014 \cite{scharstein2014high}, with varying resolution and image content.
We will focus on this latter version, namely \textit{Middlebury 2014}, since it provides an online benchmark for evaluation and still represents one of the most challenging datasets for stereo matching.

\textbf{Middlebury 2014 \cite{scharstein2014high}. } It consists of 33 scenes, divided into \textit{training}, \textit{additional} and \textit{test} splits made of respectively 13, 10 and 10 stereo pairs. Some of the data are used multiple times under different exposure and illumination conditions.  A unique feature of this dataset is the very high image resolution, which reaches 6 megapixels compared to 0.3 megapixels of the KITTI images, and a disparity range between 200 and 800 pixels, representing one of the hardest challenges of this dataset. \matteo{Images and ground truth disparity maps are provided at full (F), half (H) and quarter (Q) resolution.}
An active stereo pipeline, described in detail in \cite{scharstein2014high}, was deployed to obtain dense and accurate ground truth depth.
The limited number of training samples and the variety of content in the images make this dataset particularly challenging for deep learning methods, in particular for end-to-end models as we will set in the next sections.
\matteo{The online benchmark\footnote{\url{vision.middlebury.edu/stereo/eval3}}, reports the percentage of pixels having disparity errors larger than 0.5, 1, 2 and 4, as well as average and root mean square errors (RMSE) and other metrics
on either all or non-occluded pixels.}

\subsection{ETH3D}

ETH3D \cite{schops2017multi} is a recent, real-world, 
multi-view dataset for 3D reconstruction acquired in both indoor and outdoor environments at ETH Zurich.
It consists of 25 high-resolution color multi-view stereo scenes divided into 13 for training and 12 for testing, 10 low-resolution grayscale many-view videos evenly divided for training and testing and finally 47 low-resolution grayscale stereo pairs, respectively split into 27 and 20 for training and testing.
To obtain ground truth disparities, the authors recorded the scene geometry with a Faro Focus X 330 laser scanner, taking one or more 360$^\circ$ scans with up to 28 million points each. Together with depth, the color of each 3D point captured by the laser scanner’s integrated RGB camera was acquired, taking about 9 minutes to collect a single scan.
\matteo{The online benchmark\footnote{\url{eth3d.net/low_res_two_view}}, reports similar metrics to those of the Middlebury 2014 dataset.}

\subsection{Freiburg SceneFlow}
\label{sec:freiburg}

The Freiburg SceneFlow dataset \cite{mayer2016large,mayer2018makes} was a ground-breaking step forward in the field. As evidence, we underline that most of the proposed end-to-end networks for stereo matching are trained from scratch on this large dataset, before being fine-tuned on real data. 
The dataset consists of 3D scenes, from which images and dense ground truth for stereo, optical flow, and scene flow are rendered.
To this end, the authors modified the internal rendering engine of the freely available Blender suite in order to produce fully dense and accurate ground truth for the two views of a virtual stereo camera with a resolution of $540\times960$ pixels.
The dataset is organized into three subsets, named \textit{FlyingThings3D}, \textit{Monkaa} and \textit{Driving}, totalling about 39000 stereo pairs overall. We briefly summarize the three datasets, referring the reader to \cite{mayer2018makes} for more details.

\textbf{FlyingThings3D.} This set of images has been obtained fully automatically: the authors created a structured background from random geometric shapes, and overlayed on it dynamic foreground objects sampled from ShapeNet \cite{chang2015shapenet} and following linear trajectories in 3D space, as the camera itself does. 
It totals 22872 stereo pairs, while 4370 more are set aside as the validation set of the full SceneFlow dataset.

\textbf{Monkaa.} In contrast to FlyingThings3D, stereo pairs contained in this split are generated from an animated movie in a deterministic way. 3D artists modeled original scenes and elements, then the authors produced custom environments and rendered long scenes to sample sufficient data. This subset contains 8591 stereo pairs.

\textbf{Driving.} Similarly to Monkaa, this split has been generated in a deterministic way as well. The aim of this portion of the Freiburg dataset is to provide data relevant to driving environments, as opposed to general scenes. This set contains 4392 samples.

\subsection{Other datasets}

We mention a few more datasets which have been used less widely with respect to the previous ones.

\matteo{
\textbf{MPI-Sintel \cite{butler2012sintel}.} Originally proposed for dense optical flow benchmarking, the Sintel dataset is a collection of synthetic images extracted from short, animated movies. Together with dense flow labels, it recently has made available stereo pairs, disparity and occlusion ground truth annotations for 23 scenes with a total of 1109 stereo pairs. However, this dataset has rarely been used in the evaluation of stereo approaches \cite{Ilg_2018_ECCV,saikia2019autodispnet}; 
it is much more popular in optical flow.

\textbf{CARLA \cite{Dosovitskiy17}.} This framework implements a simulator allowing for data generation in the context of autonomous driving. It provides open-source code and digital assets (\eg urban layouts, buildings, vehicles), allowing for synthesis of virtual environments in varying weather conditions and with full control of static and dynamic actors in the scene. Agents with custom sensors, such as a stereo camera, can navigate in the simulated world and acquire a potentially unlimited amount of images with dense annotations. Although rarely used to train stereo networks \cite{Tonioni_2019_learn2adapt}, the aforementioned modelling power makes it a promising tool for future research. 


\textbf{More driving datasets.} We conclude listing newly-proposed imagery acquired in driving environments.} The Oxford Robotcar dataset \cite{RobotCarDatasetIJRR} has been acquired after more than 100 km navigation with a trinocular camera, thus collecting stereo pairs with both a narrow and a wide baseline. Ground truth depth is generated from raw LIDAR measurements. Apolloscape \cite{apolloscape} provides 5165 stereo pairs at 3 megapixels resolution, divided into 4156 pairs for training and 1009 for testing, with dense ground truth obtained by point cloud accumulation and fitting 3D CAD models, similarly to KITTI 2015. The recent DrivingStereo dataset \cite{Yang_2019_CVPR} provides over 180k stereo pairs at 1.4 megapixels resolution, with semi-dense ground truth disparities obtained by interpolating LIDAR measurements and refining them with a deep stereo network.
Potentially relevant are very large datasets released to support research on autonomous driving. Specifically, the Waymo Open Dataset \cite{waymo_open_dataset}, Argoverse by Argo AI \cite{Argoverse} and the Lyft Level 5 dataset \cite{lyft2019} are of unprecedented scale and one could imagine rectified stereo pairs with ground truth being extracted from them. We anticipate that Apolloscape and DrivingStereo, which provide binocular stereo imagery directly, will play a significant role for future developments in stereo matching.

\begin{figure*}[h]
    \centering
    \renewcommand{\tabcolsep}{1pt}
    \begin{tabular}{cccc}
        \includegraphics[width=0.25\textwidth]{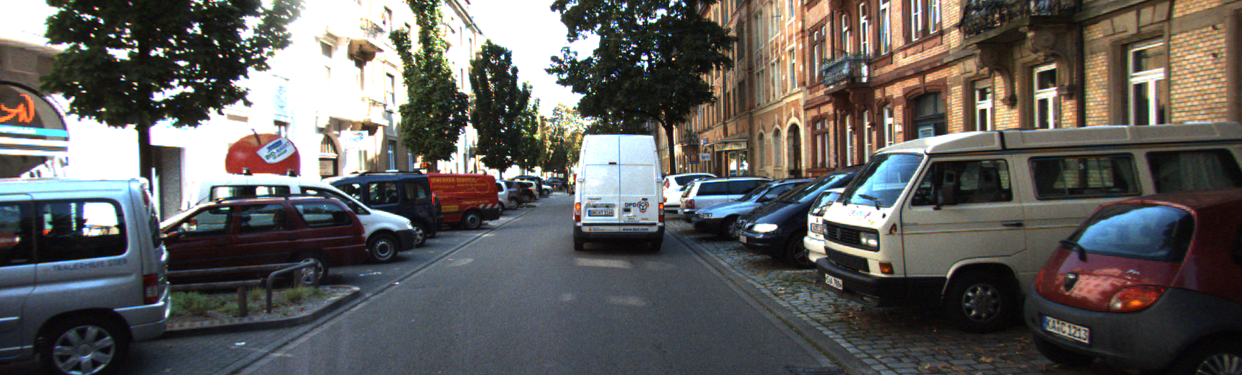} &
        \includegraphics[width=0.25\textwidth]{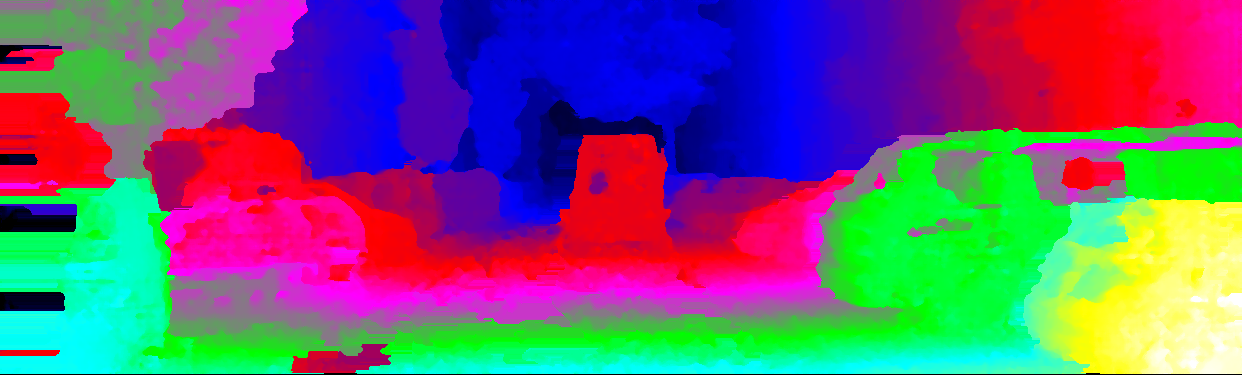} &
        \includegraphics[width=0.25\textwidth]{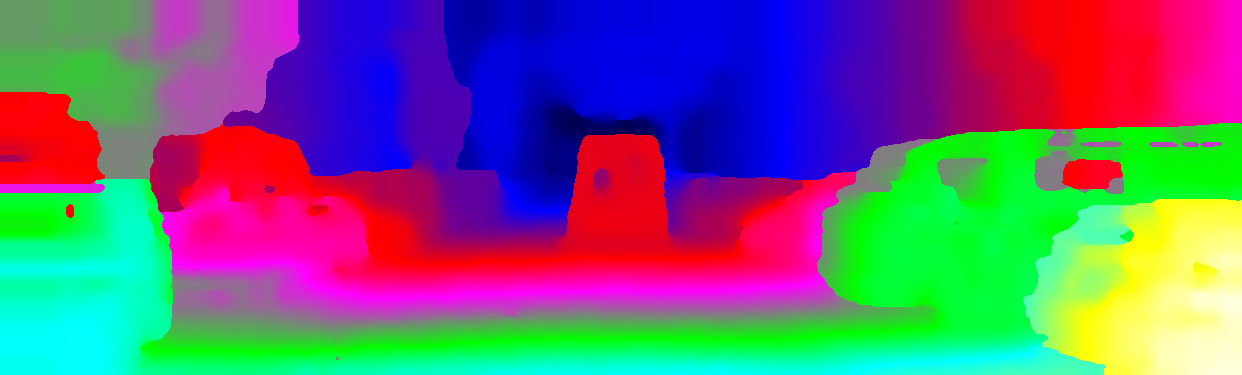} &
        \includegraphics[width=0.25\textwidth]{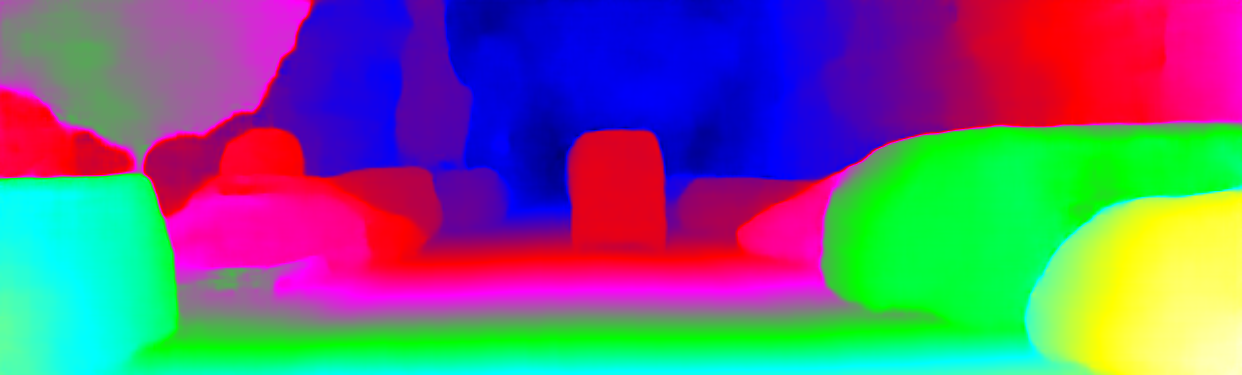} \\
    \end{tabular}
    \caption{\textbf{Evolution of stereo algorithms.} From left, reference image from KITTI 2015, disparity maps by SGM \cite{hirschmuller08}, MC-CNN-acrt \cite{zbontar2016stereo} and DispNetC \cite{mayer2016large}. Learned matching costs outperform traditional pipelines, while end-to-end models perform even better in challenging regions (\eg cars).}
    \label{fig:evolution}
\end{figure*}

\section{Learning within the stereo pipeline}
\label{sec:pipeline}

Despite the proliferation and success of deep learning \cite{lecun2015deep} in most  high-level tasks in computer vision, low-level vision problems were only partially affected at the very beginning. 

The initial research efforts applying machine learning in stereo vision aimed at improving individual steps of the established pipeline \cite{scharstein2002taxonomy}, for instance by learning a matching cost function to replace hand-crafted ones based on SAD or the census transform \cite{zabih1994non} or by learning how to improve the subsequent optimization and refinement stages after the conventional winner-takes-all (WTA) strategy.
This first step ignited the rapid evolution of stereo algorithms of the last five years, progressively developing more robust methods as shown in \autoref{fig:evolution}.

\subsection{Matching cost} 

Since stereo matching aims to detect correspondences between pixels, intuitively learning a robust matching function is a promising first step. Better matching costs also allow for improved volume optimization and thus lead to more accurate disparity maps.
Critical for this kind of approaches is the possibility of extracting large amounts of training data from a few hundred images. Since the goal is learning correspondences between pixels, each pixel with available ground truth represents a training sample. This means that a relatively small dataset, such as KITTI 2015, provides more than 30 million samples, even though only 30\% of the total pixels are labelled.

\textbf{MC-CNN \cite{zbontar2016stereo}.} The most impactful work in this area is by \v{Z}bontar and LeCun who train a CNN to predict whether two image patches match or not. 
A Siamese network extracts features from the two images, which are passed to a fully connected network estimating a matching score for the center pixel of the left patch.
By replacing fully connected layers with $1\times1$ convolutions, the architecture can be made fully convolutional to process the entire image at once. 
Two versions were developed: \textit{MC-CNN-acrt} for which the features are concatenated, thus $D$ forwards are required at test time (where $D$ is the disparity range) and \textit{MC-CNN-fst} which replaces concatenation with a dot product, allowing for a single \matteo{forward pass through the network} at the cost of a small drop in accuracy. In order to achieve state-of-the-art results, the cost volume obtained by MC-CNN is optimized and refined using a conventional SGM pipeline \cite{mei2011building} including Cross Based Cross Aggregation (CBCA) \cite{zhang2009cross}.

\textbf{Deep Embed \cite{chen2015deep}.} In conventional pipelines, the choice of window size is crucial to the effectiveness of local aggregation. In particular, large windows allow for processing more information and are more robust to textureless regions, but produce blurred boundaries near depth discontinuities. Conversely, small windows are preferred near edges but are ineffective in ambiguous regions. Following this observation, Chen \etal design a network to learn a multi-scale feature embedding, processing $13\times13$ patches at full and half resolution, thus learning a cost function from both small and large windows. Final matching scores are obtained as the dot product between left and right multi-scale features, extracted by a Siamese feature extractor. 

\textbf{Content CNN \cite{luo2016efficient}.} The aforementioned approaches 
process patches separately, producing a score for each patch comparison that is independent of other comparisons. Luo \etal pose the problem as multi-class classification, where the classes are all possible disparities, instead of binary classification for each disparity. This leads to calibrated scores for each disparity and higher accuracy. The dot product, as in MC-CNN-fst, is used to combine left and right features.

\textbf{Per-pixel pyramid-pooling \cite{park2017lwcnn}.} Park and Lee enable the network to access wider context by adding a pyramid pooling layer that considers data over multiple scales without loss of resolution and detail. This leads to disparity maps with precise discontinuities and higher accuracy than MC-CNN-acrt, especially when avoiding SGM optimization.

\textbf{SDC \cite{Schuster_2019_CVPR}.} Schuster \etal propose a novel architecture for learning an universal descriptor for dense matching. By leveraging parallel dilated convolutions, with different dilation factors, SDC extracts features by processing a large receptive field with moderate increase of the computational cost. This solution is effective at improving performance of stereo, optical flow and scene flow algorithms when replacing traditional descriptors.

\textbf{Consistency and distinctiveness \cite{zhang_tip17}.} Zhang and Wah argue that almost all existing problems in dense matching are caused by features that violate the principle of consistency, the principle of distinctiveness or both. Consistency requires that a given point should have similar descriptors when it is observed from different viewpoints. Distinctiveness states that a feature should be different from other pixels in its surrounding regions. The author seek guide features in a deep multi-objective optimization framework incorporating both principles.

\textbf{CBMV \cite{batsos2018cbmv}.} Batsos \etal propose a method for learning the matching volume leveraging both data with ground truth and conventional wisdom. A random forest classifier determines the likelihood of whether a given disparity for a pixel is correct based on a combination of hand-crafted matching functions and long-range constraints. The resulting cost volume is optimized as in \cite{zbontar2016stereo}, leading to similar accuracy when testing in the training domain, but much better generalization across different domains.

\textbf{Weakly-supervised deep metric \cite{Tulyakov_2017_ICCV}.} Learned matching functions achieve high accuracy, but require substantial amounts of annotated data for training.
Tulyakov \etal propose an effective strategy to leverage coarse information from the stereo setup, such as epipolar, uniqueness, smoothness and ordering constraints, to obtain  weak supervision from stereo pairs with spare or no ground truth. Despite the weak supervision, the learned matching functions perform as well as those trained conventionally.

\subsection{Optimization}
After initial cost volume computation, optimization is crucial for gathering information from a larger context and overcoming the limitations of pixel-wise matching.
SGM \cite{hirschmuller08} is by far the most popular conventional technique for cost volume optimization; as a result, improving it via learning has received attention from the research community.

\textbf{GCP \cite{spyropoulos2014learning,spyropoulos2016correctness}.} Based on the assumption that reliable pixels can be used to influence neighboring pixels within a global optimization framework, Spyropoulos \etal select highly reliable pixels, detected by a random forest classifier, as ground control points (GCPs). GCPs are, then, used to introduce soft constraints into the matching volume, which is optimized using MRF energy minimization \cite{komodakis_cvpr07}. 

\textbf{LevStereo \cite{park2015leveraging,park2018learning}.} Park and Yoon propose a generalized modulation strategy in order to improve the robustness and the accuracy of widely used stereo matching algorithms such as SGM. More specifically, cost curves of an initial cost volume showing evidence of low confidence values are flattened while highly confident pixels are left unchanged. This modulation scheme is effective because it enhances the importance of reliable matching costs inside the SGM aggregation step, allowing reliable pixels to guide disparity estimation for unreliable ones. 

\textbf{O1 \cite{poggi2016learning,poggi2020learning}.} By analyzing in depth the SGM algorithm and observing that the Scanline Optimization (SO) strategy causes streaking artifacts in the final disparity map, Poggi and Mattoccia propose a more effective measure computed on features extracted from the disparity map only in constant time. Specifically, the standard SO scheme is replaced with a smarter strategy that properly weights the matching costs computed for each independent path using the corresponding confidence score. This leads to visible artifacts in the disparity map being considerably alleviated. 

\textbf{PBCP \cite{seki2016patch}.} Seki and Pollefeys argue that not all pixels should be subject to the same smoothness penalties in SGM optimization. If penalties were decreased at the most confident pixels, scanline optimization would propagate information from reliable to unreliable pixels. This can be achieved by changing the  SGM formulation, by adjusting the smoothness penalty parameters per pixel according to confidence scores, estimated by a CNN processing the initial WTA left and right disparity maps.

\textbf{SGM-Net \cite{seki2017sgm-net}.} Seki and Pollefeys extend PBCP to distinguish between positive and negative disparity transitions along the scanlines, since they signify different occlusion relationships. They introduce a new loss function, that includes path and neighbor costs by taking into account the cost of the disparity path over a scanline compared to the ground truth and transitions between neighboring pixels, respectively.

\textbf{SGM-Forest \cite{schonberger2018sgm-forest}.} Following the rationale in \cite{poggi2016learning}, Sch\"onberger \etal develop a random forest classifier for improving the selection among disparity values for a pixel proposed by multiple scanlines in SGM. 
The classifier considers disparities and optimized costs per scanline to produce per-pixel scores, used both to combine the disparity hypotheses from the different scanlines, as well as to obtain confidence.

\subsection{Refinement}
The last step of the pipeline aims to refine the estimated disparity map. Traditionally, image processing techniques like median or bilateral filtering are used for this task, after a left-right consistency check.
Recently, neural networks have been proposed to replace traditional image filters. 

\textbf{GDN \cite{shaked2017improved}.} Shaked and Wolf develop a multi-stage architecture, named L-ResMatch, that addresses cost volume refinement in its last stage. L-ResMatch begins with a residual network that learns a matching cost function. Then, traditional aggregation steps like CBCA and SGM \cite{zbontar2016stereo} are applied. Finally, a Global Disparity Network (GDN) locally refines the optimized cost volume to further improve the quality of the final disparity map, while predicting a confidence estimate for each pixel at the same time.

\textbf{Detect Replace Refine (DRR) \cite{gidaris2017DRR}.} Gidaris and Komodakis present the DRR algorithm, which decomposes label improvement in a detection, a replacement and a refinement step. DRR is based on the hypothesis that hard mistakes should be detected and replaced, because correcting them does not depend on the wrong input estimates, while soft mistakes can be corrected by additive refinement. The authors show that further improvements can be achieved if the network is applied iteratively.

\textbf{Order-based Surface Decision (OSD) \cite{ye_ieee17}.} Ye \etal extend DRR \cite{gidaris2017DRR} by distinguishing among different failure modes of the matching process. Different strategies for each case, and corresponding sub-networks that implement them, are introduced. The resulting system is able to improve the outputs of a diverse set of matching algorithms on the Middlebury 2014 benchmark.

\textbf{RecResNet \cite{batsos2018recresnet}.} Batsos and Mordohai apply a dense label correction algorithm, implemented as a recurrent, residual network, to an input disparity map estimated by a black box stereo algorithm. The output disparity map is generated based on the noisy input disparity map and the left image by applying a combination of residuals computed at multiple scales, to correct heterogeneous types of errors. The same network is applied recurrently to its own output to make further improvements. 

\textbf{LRCR \cite{Jie_2018_CVPR}.} Taking advantage of disparity estimates from both the left and the right view, 
the \textit{Left-Right Comparative Recurrent} (LRCR) model embeds the left-right consistency check into a unified pipeline in order to improve the final disparity estimate. A soft attention mechanism, jointly with recurrent learning, is in charge of selecting areas in the images for refinement, thus guiding the network to correct errors mainly on unreliable initial depth estimates.  

\textbf{VN \cite{knobelreiter2019learned}.} Departing from the more common refinement processes based on residual corrections, Kn{\"o}belreiter and Pock propose a learning-based model built on a variational refinement network for the same purpose. In particular, the network takes the RGB image, the initial disparity map and a confidence measure as input, and performs collaborative denoising, considering that errors can be identified by considering the three inputs jointly.

\subsection{Experimental comparison}
\label{sec:pipeline_results}

In this section, we report a quantitative comparison between the approaches that apply learning to stages of the stereo pipeline discussed so far. To ensure a fair comparison, we retrieve results from popular online benchmarks for stereo matching \cite{Geiger2012CVPR,menze2015object,scharstein2014high,schops2017multi}. Since not all methods have been submitted to the online benchmarks, we focus on the subset providing such results. If available, we select results not labelled \textit{ROB} since  training is likely to be on the data provided by the benchmark itself.

\textbf{KITTI 2015 \cite{menze2015object}. } Table \ref{tab:kitti-pipeline} shows results of methods covered in this section on the KITTI 2015 leaderboard. \matteo{At the bottom of the table, we also report the results of SGM, as a non data-driven baseline, highlighted in yellow.} First, we can see how all methods outperform SGM by a large margin. MC-CNN achieves a major boost in accuracy, while PBCP and SGM-Net that built upon it further improve in accuracy. Unsurprisingly, disparity refinement approaches attain the best results among methods reviewed in this section since they are applied on the output of other algorithms.

\textbf{Middlebury 2014 \cite{scharstein2014high}. } Table \ref{tab:middlebury-pipeline} summarizes results from the Middlebury benchmark. In this case, we can see how SGM-forest performs much better compared to what is observed on KITTI.
Methods leveraging on deeper models, such as refinement techniques, do not appear on this online benchmark. The same is true for end-to-end models that we are going to discuss in the remainder. This is due to the small number of training images available for fine-tuning these more complex networks.

\textbf{ETH3D \cite{schops2017multi}. } We report, for the sake of completeness, results on the ETH3D in Table \ref{tab:middlebury-pipeline} although this benchmark was published later than the ones above.
SGM-forest confirms its good performance on ETH3D as well.

\begin{table}
\centering
\scalebox{0.8}
{
\centering\renewcommand{\tabcolsep}{2pt}
\begin{tabular}{| l | c | c | c | c |}
\hline
 & \multicolumn{4}{c|}{\cellcolor{blue!5}\textbf{KITTI 2015} \cite{menze2015object}} \\
 \cline{2-5}
{\bf Method} & {\bf D1-bg\%} ($\bm{\downarrow}$)& {\bf D1-fg\%} ($\bm{\downarrow}$)& {\bf D1-all\%} ($\bm{\downarrow}$)& {\bf time (s)} ($\bm{\downarrow}$)\\ \hline
LRCR \cite{jie2018LRCR} & 2.55 & 5.42 & 3.03 & 49.2 \\
RecResNet \cite{batsos2018recresnet} & 2.46 & 6.30 & 3.10 & 1.3 \\
DRR \cite{gidaris2017DRR} & 2.58 & 6.04 & 3.16 &  0.4 \\
L-ResMatch \cite{shaked2017improved} & 2.72 & 6.95 & 3.42 & 48 \\
PBCP \cite{seki2016patch} & 2.58 & 8.74 & 3.61 & 68 \\
SGM-Net \cite{seki2017sgm-net} & 2.66 & 8.64 & 3.66 & 67 \\
MC-CNN-acrt \cite{zbontar2016stereo} & 2.89& 8.88& 3.89& 67\\
SGM-Forest \cite{schonberger2018sgm-forest} & 3.11 & 10.74 & 4.38 & 6 \\
Content-CNN	\cite{luo2016efficient} & 3.73 & 8.58 & 4.54 & 1 \\
VN \cite{knobelreiter2019learned} & 4.29 & 7.65 & 4.85 & 0.5 \\
CBMV \cite{batsos2018cbmv} & 4.17 & 9.53 & 5.06 & 250 \\
\rowcolor{LightYellow}OpenCV-SGBM \cite{hirschmuller08} & 8.92 & 20.59 & 10.86 & 1.1\\
\hline
\end{tabular}
}
\smallskip
\caption{\textbf{KITTI 2015 leaderboard \cite{menze2015object}}, showing methods learning stages of the pipeline. }
\label{tab:kitti-pipeline}
\end{table}

\begin{table}[t]
\centering
\scalebox{0.8}
{
\centering\renewcommand{\tabcolsep}{2pt}
\begin{tabular}{| l | c | c | c | c | c |}
\hline
 & \multicolumn{3}{c|}{\cellcolor{blue!5}\textbf{Middlebury 2014} \cite{scharstein2014high}} & \multicolumn{2}{c|}{\cellcolor{blue!5}\textbf{ETH3D} \cite{schops2017multi}} \\
\cline{2-6}
{\bf Method} & {\bf Res.} & {\bf bad 2.0\%} ($\bm{\downarrow}$)& {\bf avg. px} ($\bm{\downarrow}$)& {\bf bad 1.0\%} ($\bm{\downarrow}$)& {\bf avg. px} ($\bm{\downarrow}$)\\ \hline
SGM-Forest \cite{schonberger2018sgm-forest} & H & 7.37 & 2.84 & 4.96 & 0.36 \\
MC-CNN-acrt \cite{zbontar2016stereo} & H & 8.08	 & 3.82 & - & - \\
CBMV \cite{batsos2018cbmv} & H & 11.1 & 4.71 & 5.35 & 0.33 \\
VN \cite{knobelreiter2019learned} & H & 14.2 & 2.49 & - & - \\
\rowcolor{LightYellow} SGM \cite{hirschmuller08} & H & 18.4 & 5.32 & 10.08 & 0.50 \\
\hline
\end{tabular}
}
\smallskip
\caption{\textbf{Middlebury 2014 \cite{scharstein2014high} and ETH3D \cite{schops2017multi} leaderboards}, showing methods learning stages of the pipeline. }
\label{tab:middlebury-pipeline}
\end{table}

\section{End-to-end deep learning}
\label{sec:e2e}

Although machine learning substantially improved each step of the traditional stereo matching pipeline \cite{scharstein2002taxonomy}, the introduction of end-to-end models drove the community towards a new paradigm.
As can be seen in popular benchmarks, KITTI 2012 \cite{Geiger2012CVPR} and 2015 \cite{menze2015object}, in just a few years, end-to-end methods dominated dense disparity estimation, thanks to the availability of large amounts of labeled data. \matteo{Indeed, while few hundred annotated images \cite{Geiger2012CVPR,menze2015object,scharstein2014high} are enough to train learning-based pipeline stages, they are not for end-to-end models. However, thousands of labeled pairs can be made available for free using graphics \cite{mayer2016large,mayer2018makes}, overcoming the costly and cumbersome process of labeling a large number of images with accurate depth measurements.

The rest of this section is organized as follows: \autoref{sec:taxonomy} introduces a taxonomy of deep models for end-to-end disparity estimation, divided in 2D and 3D architectures.  We review the most relevant models for both classes in \autoref{sec:2d} and \autoref{sec:3d}, respectively. 
Finally, \autoref{sec:results-endtoend} summarizes the performance of these networks on the benchmarks.}

\subsection{Taxonomy}
\label{sec:taxonomy}

According to the popular online benchmarks \cite{Geiger2012CVPR,menze2015object,scharstein2014high,schops2017multi}, there are hundreds of published, and unpublished, deep networks competing for the top ranks. We can broadly categorize them into two distinct classes according to their design: \textit{2D architectures} and \textit{3D architectures}.
\matteo{The main difference between the two is the strategy deployed to encode features and geometry. 
Next, we discuss the differences between the two categories, while introducing models from the literature belonging to both, highlighting their distinctive features and the rationale behind their design.} For additional details, such as training schedules or layer configuration, we refer readers to the corresponding papers. 

\subsection{2D architectures} 
\label{sec:2d}

This family of deep networks is closer to neural models designed to solve other dense regression tasks such as semantic segmentation \cite{deeplabv3plus2018}, optical flow \cite{dosovitskiy2015flownet} or monocular depth estimation \cite{fu2018supervised}. These architectures usually deploy an encoder-decoder design, inspired by the U-Net model \cite{ronneberger2015u} to keep memory requirements and runtime manageable as well as to increase the receptive field of the network to leverage image context. Thanks to the efficiency of 2D convolution operations on modern GPUs, some of these models achieve from a few to dozens of frames per second at the cost of a negligible loss of accuracy \cite{yin2019hierarchical,Tonioni_2019_CVPR}. Pivotal to the spread of these architectures is the work by Mayer \etal \cite{mayer2016large} that introduced a custom layer, namely the \textit{correlation layer}, in charge of computing similarity scores between features extracted from the two images. 

\textbf{DispNet-C \cite{mayer2016large}.} 
The work by Mayer \etal \cite{mayer2016large} is a milestone for the switch to end-to-end disparity regression. Following the U-Net design \cite{ronneberger2015u}, the authors propose DispNet-S, an encoder-decoder architecture. The first portion of the architecture feeds the input images to several 2D convolutional layers that decimate the input resolution. 
Then, stacks of 2D deconvolution layers gradually restore the resolution up to half the original, matching the full resolution using bilinear upsampling. In order to preserve fine details that were lost during downsampling, features from the encoding module are concatenated to corresponding ones at the same resolution extracted by the decoder.
Due to this design choice, the network is large enough to learn disparity inference but is not trained to reason about correspondences explicitly. This latter behavior is attained by introducing a \textit{correlation layer}, previously proposed in \cite{dosovitskiy2015flownet}, computing similarity between patches $x_1$ and $x_2$ on the two images:

\begin{equation}
    c(x_1, x_2) = \sum_{o\in[-k,k]\times[-k,k]} \langle f_1(x_1 + o),f_2(x_2 + o) \rangle
\end{equation}
with the two patches of size $K=2k+1$. 
Despite the general formulation, $K$ is typically set to 0, thus computing single-pixel correlations.
A siamese sub-network extracts features from both images at the beginning.
Then, correlating each patch in the reference image and all candidates in a 
$(2D+1)$ search range, yields an equal number of correlation maps, stacked along the channels dimension and forwarded to subsequent 2D convolutional layers.
According to \cite{mayer2016large}, introducing the correlation layer significantly improves the quality of estimated disparity maps, in particular on real datasets such as KITTI 2012 and 2015.

\textbf{CNN+CRF \cite{Knobelreiter_2017_CVPR}.} Although not strictly an end-to-end network, we include the work by Kn\"obelreiter \etal here.
The authors design a joint CNN and Conditional Random Field (CRF) model to infer dense disparity maps. Joint training is made possible by formulating the CRF as a maximum margin Markov network. We consider this as an interesting intermediate step between models aimed at implementing single steps in the pipeline and a fully end-to-end CNN.

\textbf{CRL \cite{pang2017cascade}.} Although DispNet-C fails to outperform conventional hand-engineered stereo pipelines such as \cite{zbontar2016stereo}, it paved the way for the deployment of more sophisticated deep models.
The first 2D architecture to rise in the KITTI leaderboards is the one by Pang \etal. Starting from the observation that learning a residual signal \cite{he2016deep} is usually easy for a neural network, the authors combine a DispNet-C model with a second subnetwork, referred to as DispResNet, which is in charge of computing residual corrections to the initial disparity map, $d_1$, estimated by the DispNet-C subnetwork. To do so, $d_1$ is used to warp the right input image towards the left. The left image, the warped right image and their difference are fed to the DispResNet module which determines the adjustment to $d_1$.

\textbf{iResNet \cite{liang2018learning_1st_Rob}.} 
Liang \etal develop a deep network, inspired by the different steps in the conventional stereo pipeline. The first portion extracts features at multiple scales and feeds them to a correlation layer, producing an initial disparity estimate. Then, as in CRL, this estimate is used to warp right-image features to the left camera, before being passed to a refinement module made of a new correlation layer with smaller search range and additional 2D convolution and deconvolution operators. This module is stacked multiple times for iterative refinement, and the architecture is dubbed the Iterative Residual Network (iResNet). iResNet ranked first in the 2018 Robust Vision Challenge.

\textbf{DispNet-CSS \cite{Ilg_2018_ECCV}.} Following the successful findings about residual learning, Ilg \etal address occlusions, motion and depth boundary estimation. 
Specifically, a stack of networks iteratively produces improved predictions, which are followed by subsequent residual refinements. Either a DispNet-CSS or a FlowNet-CSS can be trained for stereo matching or optical flow estimation, respectively, based on this scheme. In both cases, only the first subnetwork deploys a correlation layer. Combining the two CSS architectures, the authors also estimate scene flow. 

\textbf{AutoDispNet-CSS \cite{saikia2019autodispnet}.} While a general-purpose encoder-decoder can perform reasonably well if trained for a specific task, the literature (and this survey as well) supports the thesis that careful design choices and tuning are necessary to obtain a state-of-the-art model.
These can be accomplished automatically using \textit{auto machine learning} (autoML) to optimize for the best stereo matching architecture.
By defining a set of candidate operations, \eg 2D convolutions and upsampling layers, Saikia \etal use the gradient-based method DARTS \cite{liu2018darts} to explore the space of architectures and find the best configuration for DispNet. 
Despite deploying only a subset of the possible design choices among the candidate layers, the obtained AutoDispNet-CSS performs comparably to state-of-the-art.

\textbf{MADNet \cite{Tonioni_2019_CVPR}.} Although most of the architectures focus on accuracy, efficiency is also crucial, in particular in real-world applications. 
Tonioni \etal apply a coarse-to-fine strategy within their \textit{Modularly Adaptive Network} (MADNet). 
Starting from the coarser level of a feature pyramid extracted from each image, features are fed to a correlation layer, then an initial disparity map is predicted by a 2D convolutional decoder and upsampled to the previous level of the pyramid. This predicted disparity is used to warp the right features to the left view, feeding them to another correlation layer with a small search range to estimate a better disparity map. 
The procedure continues until the highest resolution, where a refinement network with 2D dilated convolutions produces the final disparity map. 

\textbf{HD$^3$ \cite{yin2019hierarchical}.} Along these lines, a coarse-to-fine strategy has been exploited by other authors. Yin \etal couple it with discrete distribution estimation, which is used to estimate the degree of uncertainty of the predicted values, an aspect typically neglected by the regression approaches discussed so far. 
Following a pyramidal approach similar to \cite{Tonioni_2019_CVPR}, a correlation layer is fed with left and warped right features and a decoder extracts features in the density embedding space in order to predict a final match density. The conversion between dense distributions and motion fields (either flow or disparities) and vice-versa is necessary to handle upsampling and to generate supervision for the next level.

\textbf{SegStereo \cite{yang2018segstereo}.} 
Multi-task learning \cite{ruder2017overview} has gained popularity in various areas of computer vision, where joint learning of multiple tasks leads to overall improvement in all tasks.  
Following this rationale, Yang \etal propose SegStereo for joint disparity estimation and semantic segmentation. The network extracts a representation with shared features for both tasks. The features are used as input to a correlation layer and a segmentation subnetwork. The output correlation maps and semantic embeddings are concatenated and forwarded to an encoder-decoder producing the disparity map. Semantic masks are obtained for both the left and right views and forced to be consistent by warping them according to the predicted disparity. Tackled together, both tasks achieve improved results.

\textbf{EdgeStereo \cite{song2018edgestereo,song2020edgestereo}.} Depth discontinuities represent some of the most challenging regions in the image for stereo matching. Evidence for this can be observed by looking at qualitative examples of disparity maps from the KITTI online benchmarks \cite{Geiger2012CVPR,menze2015object}. Motivated by this evidence, Song \etal jointly learn disparity estimation and edge segmentation within their EdgeStereo framework. An edge detection subnetwork extracts a set of features used to estimate an edge map. They are processed together with conventional correlation scores to obtain the final disparity map, improving accuracy near depth discontinuities. 

\textbf{DSNet \cite{zhan2019dsnet}.} Following the multi-task trend, Zhan \etal propose a network with an encoder which is shared between semantic segmentation and disparity estimation, and is trained in a multi-stage manner. At first, it is optimized for semantic segmentation; then, the weights are fixed and used to learn disparity estimation by means of a matching module combining feature correlation and concatenation by means of attention mechanisms. Finally, a third stages optimizes the model for both tasks jointly.

\textbf{SENSE \cite{Jiang_2019_ICCV}.} Another step in the direction of multi-task learning is performed by Jiang \etal, proposing a single, compact architecture estimating disparity, optical flow, disparity change and semantic segmentation at once. Starting from a single, shared encoder, different decoders are introduced for each task.

\textbf{Unsupervised Stereo \cite{zhou2017stereo}.} Although they achieve compelling results, deep networks are heavily dependant on the amount and variety of the labeled training data.  
To simplify the training process of deep stereo networks, Zhou \etal propose an end-to-end framework capable of learning disparity estimation in an unsupervised manner. A 2D network with an image-guided aggregation network is designed to estimate disparity maps for the left and right images. Then, left-right consistent matches are used to train the network iteratively on its own predictions.
Moreover, the proposed architecture has accuracy comparable to DispNet-C, when trained with supervision.

\subsection{3D architectures} 
\label{sec:3d}

Whereas 2D networks are much closer to traditional neural models, 3D architectures were developed specifically for stereo matching. Although the traditional encoder-decoder design is embodied by these frameworks as well, they differ from the previous category by explicitly encoding geometry during the processing of the features. \matteo{Conversely to 2D models, 3D networks explicitly encode matching properties between pixels in the form of feature vectors utilizing different operators, \eg concatenation and feature difference. By performing this operation on the entire search range $D$, 3D architectures produce an output volume of increased dimensionality: $D\times H \times W$, times the amount of features F, resulting in a 4D data structure. Subsequently, the 4D tensor is processed by 3D convolutions, resulting in explicit processing of a matching volume-like representation.}
This strategy comes at the cost of a much higher memory requirements and runtime.

\textbf{GC-Net \cite{kendall2017-gcnet}.} This framework represents the first attempt to deploy explicit knowledge about geometry to design a 3D neural network for stereo matching. It also was the first end-to-end model to outperform hand-crafted pipelines on the KITTI benchmarks. 
High-level features are extracted from both images using two encoders with shared weights. 
In this phase, the original resolution is halved to reduce memory requirements and then a cost volume is built by concatenating per-pixel features $F$ across the two images on the entire disparity search range, producing a $\frac{D}{2} \times \frac{W}{2} \times \frac{H}{2} \times 2F$ volume. Then, a 3D encoder-decoder module processes the volume to obtain a final $D \times H \times W \times 1$ volume, from which the disparity map is obtained using the \textit{soft-argmin} operator

\begin{equation}
  \text{soft-argmin} = \sum_{d=0}^{D} d \cdot \sigma(-c_d)  
\end{equation}
with $\sigma$ the softmax operator applied to each final feature $c_d$ along the $D$ dimension.

\textbf{ECA \cite{yu2018deep}.} Following this successful, new design paradigm, several authors focused on further boosting the accuracy of 3D networks. 
Yu \etal propose to improve the 3D optimization phase by introducing \textit{Explicit Cost Aggregation} (ECA) modules along the three different dimensions. This goal was achieved by adding a set of 3D convolutions having rectangular filters, with kernel size equal to 1 on all but one dimensions, keeping a low computational cost compared to traditional 3D convolutions. A further guided aggregation strategy is proposed, directly learning from the image a set of guides to be applied to the final cost volume before the softargmin selection.

\textbf{PSMNet \cite{chang2018psmnet}.} Although an encoder-decoder structure allows for taking into account large context information during the learning process, in 3D networks this stage occurs only after the cost volume computation, which depends on very local features. Advances in deep learning introduced new layers capable of greatly enlarging the receptive field of a neural network with a negligible computational cost, as in the case of \textit{Spatial Pyramidal Pooling layers} (SPP) \cite{he2015spatial} adopted by Chang and Chen in their \textit{Pyramidal Stereo Matching network} (PSMNet). Integrating SPP layers in the GC-Net feature extractor, together with deploying a stack of multiple 3D encoder-decoder modules proved to be effective at improving accuracy. In order to keep computational costs manageable, features are extracted down to quarter resolution before building the cost volume, thus leading to about twice as fast inference compared to GC-Net \cite{kendall2017-gcnet}.

\textbf{EMCUA \cite{Nie_2019_CVPR}.} Other authors worked on expanding the contextual information processed by neural networks in the early stages. 
Nie \etal introduce \textit{Multi-level Context Ultra-Aggregation} (MCUA). Given a branch working at a certain resolution (\ie half) in a dense network, a child module, sharing weights with the main branch, is deployed to process a downsampled version of the same features (\ie at quarter resolution). Usually, the output of the main branch reaches the same lower resolution and is processed by a new branch. At this point, features extracted by each layer of the child module are densely connected to this new branch, actually implementing inter-level interactions. 

\textbf{CSPN \cite{cheng2019spn}}
To further improve PSMNet performance, Cheng \etal \cite{cheng2019spn} propose Convolutional Spatial Propagation Network (CSPN) modules, capable of learning an affinity matrix for feature aggregation and spatial propagation of 2D unary features. By extending CSPN design to 3D, information is also propagated within the disparity dimension, enabling aggregation over both spatial and cost dimensions when processing features from 3D encoder-decoders. 

\textbf{GA-Net \cite{zhang2019ga}.} Matching cost aggregation is crucial in conventional methods, where local aggregation techniques \cite{de2011linear,hosni2012fast} or semi-global optimization \cite{hirschmuller08} are widely adopted.  Cost aggregation is beneficial even in deep neural networks, as demonstrated by the efforts to improve the design of encoder-decoder modules. Zhang \etal propose two novel layers, aimed at capturing local and global cost relationships. They are a locally guided aggregation layer and a semi-global aggregation layer, respectively implementing a traditional cost filtering strategy and a differentiable approximation of the SGM algorithm. By replacing 3D convolutions with few instances of these layers, their \textit{Guided Aggregation network} (GA-Net) easily outperforms models deploying dozens of traditional, costly convolutions.

\textbf{StereoDRNet \cite{Chabra_2019_CVPR}.} Although accurate, deep stereo networks often produce geometrically inconsistent disparity maps, which negatively affect higher-level applications such as 3D reconstruction via \textit{Truncated Surface Distance Function} (TSDF) fusion.
This leads Chabra \etal to improve PSMNet design 
\cite{chang2018psmnet} to obtain more geometrically consistent predictions and thus better 3D reconstructions by fusing them. Their contributions include the use of a Vortex pooling layer \cite{xie2018vortex} which proved to be more effective compared to the SPP layer, the introduction of 3D dilated convolutions inside the stacked encoder-decoders, and a refinement network for enhancing the initial disparity map. 

\textbf{PDSNet \cite{tulyakov2018practical}.} 
Most deep networks are memory-hungry and have to be trained for a given target disparity range.
Tulyakov \etal propose \textit{Practical Deep Stereo Network} (PDSNet) to address both limitations. They decrease the memory footprint by introducing a bottleneck matching module, which compresses the concatenated features from the two images into compact matching signatures, processed by a 3D encoder-decoder network to infer a sub-pixel MAP approximation. Thus, a weighted mean is computed around the disparity with the maximum posterior, which is robust to erroneous modes in the disparity distribution and allows to modify the disparity range without re-training. 
A novel sub-pixel criterion, derived by combining the standard cross-entropy loss with kernel interpolation, leads to faster convergence rates and higher accuracy. 

\textbf{StereoNet \cite{khamis2018stereonet}.} A weakness of 3D architectures compared to 2D models is the computational effort required even for a single inference. On average, 3D networks have about one order of magnitude higher memory requirements and runtime because of the additional dimension. The most expensive operations are those performed at the highest resolutions, thus Khamis \etal design a 3D model limited to a low-resolution volume (\ie $\frac{1}{8}$), from which a coarse disparity map is extracted. This latter is sequentially upsampled and refined through shallow 2D networks. Thanks to the much lower complexity of 2D convolutions, StereoNet achieves much higher throughput than 3D networks at the cost of a marginal accuracy drop.

\textbf{AnyNet \cite{wang2019anytime}.} 
Coarse-to-fine strategies that proved to be successful for 2D architectures have also been proposed for 3D networks. Concurrently to works on 2D networks \cite{Tonioni_2019_CVPR,yin2019hierarchical}, Wang \etal deploy a pyramidal model that extracts a small number of feature maps from the images and then builds a very compact 4D volume by computing the $L_1$ distance between left and (warped) right features. The network works at three scales, deploying a coarse to fine disparity estimation strategy. After the last prediction, a Spatial Propagation network (SPNet \cite{liu2017learning}) produces the final output. 
The authors also endow their AnyNet model with an early-stopping functionality at \textit{anytime}, \ie inference can be shortened to obtain one of the coarser disparity maps, allowing for speed-accuracy trade-offs, as in the case of real-world applications with limited resources.

\textbf{HSM \cite{yang2019hsm}.} High-resolution images have always been challenging for stereo matching, in particular in terms of resource requirements. 
As we observed so far, a standard strategy is to reduce the target resolution and rely on upsampling to restore it after inference. High-resolution stereo matching has been tackled by Yang \etal with the \textit{Hierarchical Stereo} Matching (HSM) network. Following the pyramidal approaches discussed above, they extract a set of features at different resolutions and compute cost volumes with different search ranges according to the resolution. 
Each volume is processed to obtain a disparity map, upsampled to be concatenated with higher resolution volumes, and processed to produce finer disparity maps. For training, the authors propose a new, high resolution set of images (about $2056\times2464$) combined with available high \cite{scharstein2014high} and low-resolution \cite{Geiger2012CVPR,menze2015object,schops2017multi} datasets. Moreover, they discuss the possibility of anytime on-demand inference, as in \cite{wang2019anytime}. 

\begin{table}[t]
\centering
\scalebox{0.8}
{
\centering\renewcommand{\tabcolsep}{2pt}
\begin{tabular}{| l | c | c | c | c | c |} 
\hline
 & & \multicolumn{4}{c|}{\cellcolor{blue!5}\textbf{KITTI 2015} \cite{menze2015object}} \\
 \cline{3-6}
{\bf Method} & {\bf Family} & {\bf D1-bg\%} ($\bm{\downarrow}$) & {\bf D1-fg\%} ($\bm{\downarrow}$) & {\bf D1-all\%} ($\bm{\downarrow}$) & {\bf time (s)} \\ \hline
CSPN \cite{cheng2019spn} & 3D & 1.51 & 2.88 &  1.74 & 1.0 \\
GA-Net \cite{zhang2019ga} & 3D & 1.48 & 3.46 & 1.81 & 1.8 \\
HD$^3$-Stereo \cite{yin2019hierarchical} & 2D & 1.70& 3.63& 2.02& 0.14 \\
EMCUA \cite{Nie_2019_CVPR} & 3D & 1.66& 4.27& 2.09& 0.90\\
GWC-Net \cite{guo2019group} & 3D & 1.74& 3.93& 2.11& 0.32\\
SSPCV-Net \cite{Wu_2019_ICCV} & 3D & 1.75 & 3.89 & 2.11 & 0.9 \\
HSM \cite{yang2019hsm} & 3D & 1.80& 3.85& 2.14& 0.14\\
DeepPruner \cite{Duggal_2019_ICCV} & 3D & 1.87 & 3.56 & 2.15 & 0.18 \\
DispNet-CSS \cite{Ilg_2018_ECCV} & 2D & 1.92& 3.32& 2.16& 0.25\\
AutoDispNet-CSS \cite{saikia2019autodispnet} & 2D & 1.94& 3.37 & 2.18& 0.90\\
SENSE \cite{Jiang_2019_ICCV} & 2D & 2.07 & 3.01 & 2.22 & 0.32 \\
SegStereo \cite{yang2018segstereo} & 2D & 1.88& 4.07& 2.25& 0.60\\
StereoDRNet \cite{Chabra_2019_CVPR} & 3D & 1.72& 4.95& 2.26& 0.23\\
PSMNet \cite{chang2018psmnet} & 3D & 1.86& 4.62& 2.32& 0.41\\
ECA \cite{yu2018deep} & 3D & 2.14& 3.45& 2.36& 0.22 \\
iResNet \cite{liang2018learning_1st_Rob} & 2D & 2.25& 3.40& 2.44& 0.27\\
PDSNet \cite{tulyakov2018practical} & 3D & 2.29 & 4.05 & 2.58 & 0.50 \\
EdgeStereo \cite{song2018edgestereo} & 2D & 2.27& 4.18& 2.59& 0.27\\
CRL \cite{pang2017cascade} & 2D & 2.48& 3.59& 2.67& 0.47\\
GC-Net \cite{kendall2017-gcnet} & 3D & 2.21& 6.16& 2.87& 0.90\\
CNN+CRF \cite{Knobelreiter_2017_CVPR} & 2D+CRF & - & - & 3.61 & 1.3 \\
\rowcolor{LightYellow} MC-CNN-acrt \cite{zbontar2016stereo} & - & 2.89& 8.88& 3.89& 67\\
DispNet-C \cite{mayer2016large} & 2D & 4.32& 4.41& 4.34& 0.06\\
MADNet \cite{Tonioni_2019_CVPR} & 2D & 3.75& 9.20& 4.66& 0.02 \\
StereoNet \cite{khamis2018stereonet} & 3D & 4.30& 7.45& 4.83& 0.02 \\
OASM-Net \cite{li2018occlusion} & 3D & 6.89 & 19.42 & 8.98 & 0.73 \\
\rowcolor{LightYellow} OpenCV-SGBM \cite{hirschmuller08} & - & 8.92 & 20.59 & 10.86 & 1.1\\
\hline
\end{tabular}
}
\smallskip
\caption{\textbf{KITTI 2015 leaderboard \cite{menze2015object}}, showing end-to-end methods. }
\label{tab:kitti-e2e}
\end{table}

\textbf{GWC-Net \cite{guo2019group}.} 
Various approaches for building volumes for 3D networks have been proposed, including feature concatenation, $L_1$ or $L_2$ distance. 
Guo \etal propose a new operator, namely \textit{Group Wise Correlation layer} (GWC), placed in between feature concatenation and vector correlation.
By treating the $F$ unary features as $N$ groups of structured vectors, the group-wise correlation layer computes $N$ correlation scores, producing as output a new feature vector of dimension $N$. 
The fact that $N$ is strictly smaller than $F$ by definition reduces the computational efforts required by the first 3D convolutions, which are the most expensive ones due to the higher resolution. Moreover, this scheme provides a better feature representation enabling the network to infer more accurate disparity maps.

\textbf{DeepPruner \cite{Duggal_2019_ICCV}.} Another strategy for reducing the computational burden in 3D networks is to compute the matching costs for only a subset of all possible disparity hypotheses. To this aim, Duggal \etal deploy the PatchMatch algorithm \cite{Barnes2009PAR} unrolled as a recurrent neural network to iteratively select random candidates, propagate them locally, and keep only the most promising ones. Moreover, a confidence score, which is inversely proportional to the range between the minimum and maximum disparity candidates, can be obtained for each pixel.

\textbf{SSPCV-Net \cite{Wu_2019_ICCV}.} As proved by 2D networks, \cite{yang2018segstereo,zhan2019dsnet}, joint reasoning about disparity and semantics is beneficial to both.
Along these lines, Wu \etal design a 3D network to pursue both tasks by leveraging on pyramidal cost volumes that are subsequently summed up at each resolution from coarse to fine.
Moreover, from features relevant to segmentation a semantic cost volume is built and combined with the spatial volume to further include semantic information during disparity regression.

\textbf{OASM-Net \cite{li2018occlusion}.} 
Although unsupervised learning proved to be effective thanks to image reprojection, its major shortcoming is at occlusions, where the reprojection fails to provide meaningful supervision. Li \etal address this problem by training a 3D network to jointly infer disparity and occlusions from a stereo pair. In addition to traditional reprojection losses \cite{godard2017unsupervised}, a regularization term allows for effective segmentation of occlusions, in order to properly take advantage of this information when computing the reprojection signals.

\begin{table}[t]
\centering
\centering\renewcommand{\tabcolsep}{2pt}
\scalebox{0.8}
{
\begin{tabular}{| l | c | c | c | c | c | c |}
\hline
 & & \multicolumn{3}{c|}{\cellcolor{blue!5}\textbf{Middlebury 2014} \cite{scharstein2014high}} & \multicolumn{2}{c|}{\cellcolor{blue!5}\textbf{ETH3D} \cite{schops2017multi}} \\
\cline{3-7}
{\bf Method} & {\bf Family} & {\bf Res.} & {\bf bad 2.0\%}($\bm{\downarrow}$) & {\bf avg. px} ($\bm{\downarrow}$) & {\bf bad 1.0\%} ($\bm{\downarrow}$) & {\bf avg. px} ($\bm{\downarrow}$) \\ \hline
\rowcolor{LightYellow} MC-CNN-acrt \cite{zbontar2016stereo} & - & H & 8.08 & 3.82 & - & - \\
HSM \cite{yang2019hsm} & 3D & F & 10.2 & 2.07 & 4.00& 0.28 \\
CNN+CRF \cite{Knobelreiter_2017_CVPR} & 2D+CRF & H & 12.5 & - & - & - \\

\rowcolor{LightYellow} SGM \cite{hirschmuller08} & - & H & 18.4 & 5.32 & 10.08 & 0.50 \\
EdgeStereo \cite{song2018edgestereo} & 2D & F & 18.7 & 2.68 & - & - \\
DispNet-CSS \cite{Ilg_2018_ECCV} & 2D & H & 22.8 & 4.04 & 2.69& 0.22 \\
iResNet \cite{liang2018learning_1st_Rob} & 2D & H & 22.9 & 3.31 & 3.68& 0.24 \\
StereoDRNet \cite{Chabra_2019_CVPR} & 3D & - & - & - & 4.46& 0.27 \\
DeepPruner \cite{Duggal_2019_ICCV} & 3D & Q & 30.1 & 4.80 & 3.52 & 0.26 \\
PSMNet \cite{chang2018psmnet} & 3D & Q & 42.1 & 6.68 & 5.02& 0.33 \\
\hline
\end{tabular}
}
\smallskip
\caption{\textbf{Middlebury 2014 \cite{scharstein2014high} and ETH3D \cite{schops2017multi} leaderboards}, showing end-to-end methods. }
\label{tab:middlebury-e2e}
\end{table}

\subsection{Experimental comparison}
\label{sec:results-endtoend}

In this section, we report a quantitative comparison between the end-to-end architectures discussed so far. At first, we point out that each framework was trained according to different protocols (\eg number of iterations, learning rate schedules, etc.). 
Thus, an entirely fair comparison is not possible without using the same protocol for each network.
Nonetheless, we believe that we can fairly compare networks based on their top-performing configuration reported on the leaderboards. 

\textbf{KITTI 2015 \cite{menze2015object}. } The KITTI dataset is the preferred benchmark for deep stereo models, thanks to the low-variability image content of the driving scenario and a large number of training samples available for fine-tuning. This fact becomes apparent after observing the number of deep learning entries in the leaderboard: at the time of writing more than 100 top entries make reference to deep architectures, published or unpublished.
Indeed, all the methods discussed in \autoref{sec:2d} and \autoref{sec:3d} are available on the online benchmark, with only a few exceptions \cite{zhou2017stereo,wang2019anytime,li2018occlusion,zhan2019dsnet}. \autoref{tab:kitti-e2e} collects them for a direct comparison, \matteo{reporting also results for SGM and MC-CNN-acrt as baselines (highlighted in yellow).}
As in \autoref{sec:pipeline_results}, we show D1-bg, D1-fg, and D1-all metrics together with the runtime reported by the authors, although measured on different hardware.

We observe that faster models (DispNet-C, MADNet and StereoNet) aiming for real-time performance achieve the worse D1-all score among all methods including the MC-CNN-acrt pipeline. We also point out that unsupervised models like OASM-Net already outperform conventional non-data-driven algorithms like SGM. The first end-to-end method outperforming MC-CNN-acrt is by Kn\"obelreiter \etal \cite{Knobelreiter_2017_CVPR}, combining a CNN and a CRF. Nevertheless, almost all end-to-end CNNs outperform this hybrid strategy.
\matteo{Climbing the leaderboard, we can observe how most of 3D architectures consistently outperform 2D stereo networks, with few exceptions \cite{yin2019hierarchical}, hinting that the former family excels at modeling geometry}. Nevertheless, we also observe that early 3D models were particularly slow, \eg GC-Net \cite{kendall2017-gcnet}, while the latest methods from this category are much faster, as in the case of GWC-Net \cite{guo2019group}. CSPN \cite{cheng2019spn} is the top-performing, published model on KITTI.
Overall, we believe that HSM and HD$^3$ represent the best trade-off between accuracy and speed.

\textbf{Middlebury 2014 \cite{scharstein2014high}.}
Even though end-to-end models excel on the KITTI benchmark yielding small error rates, we are far from considering stereo as a solved problem. In particular, as clearly demonstrated by the Middlebury 2014 benchmark, high-resolution images depicting heterogeneous indoor environments and the meager amount of available training samples pose a major challenge for most neural networks. Therefore, this dataset is less popular for evaluating new architectures, indeed only five of the frameworks discussed in this section appear in the leaderboard, and they are not near the top.

\autoref{tab:middlebury-e2e} collects these results. We can notice how, in general, the percentage of bad pixels is much higher compared to KITTI, with average pixel errors higher than the prefixed threshold. The much higher resolution ($\sim20\times$ compared to KITTI) and complexity of the scenes play a crucial role in making this dataset much more challenging for end-to-end models. Moreover, most methods cannot afford to process full resolution images during inference, with EdgeStereo and HSM the only models somehow dealing with them. In particular, the latter achieves the best results among all architectures, confirming the effectiveness of the hierarchical approach for high-resolution images. Nevertheless, it ranks only 31$^{st}$ on the online leaderboard. This fact highlights that end-to-end models are still far from being state-of-the-art for stereo in any environment: \matteo{indeed, the MC-CNN-acrt pipeline \cite{zbontar2016stereo} outperforms most end-to-end models, except HSM on average error.} This emphasizes how future research should focus on the development of robust networks better generalizing over content and resolution. 


\textbf{ETH3D \cite{schops2017multi}. }
\autoref{tab:middlebury-e2e} also collects the submissions to the ETH3D online benchmark by end-to-end approaches. Despite the limited training samples available, given the less heterogeneous image content and the much lower image resolution compared to the Middlebury 2014 dataset, the error rates achieved by the five architectures appearing in the leaderboard are much lower, with DispNet-CSS being the top-performing method among all submissions to the benchmark. 
This outcome confirms how most of the open challenges for deep stereo models are linked to high-resolution images together with complexity and variety of the observed environment.

\begin{figure*}[t]
    \centering
    \renewcommand{\tabcolsep}{1pt}
    \begin{tabular}{ccc}
        \includegraphics[width=0.3\textwidth]{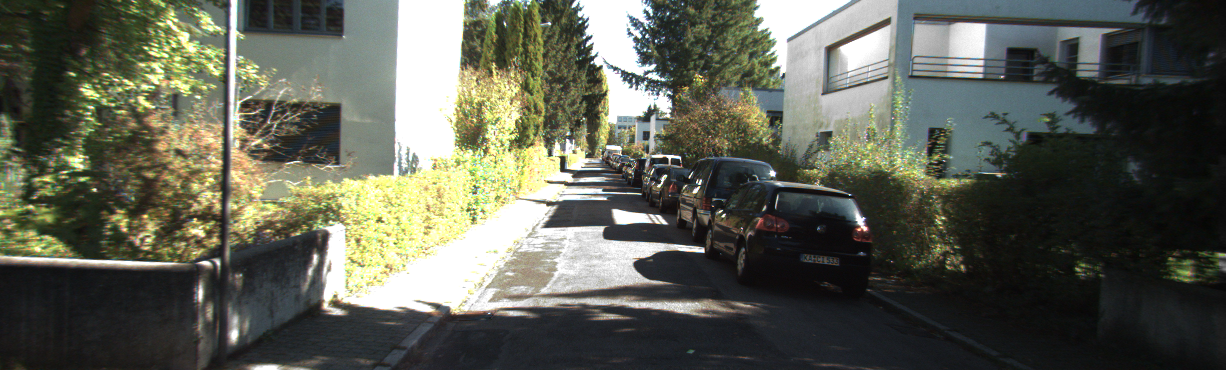} &
        \includegraphics[width=0.3\textwidth]{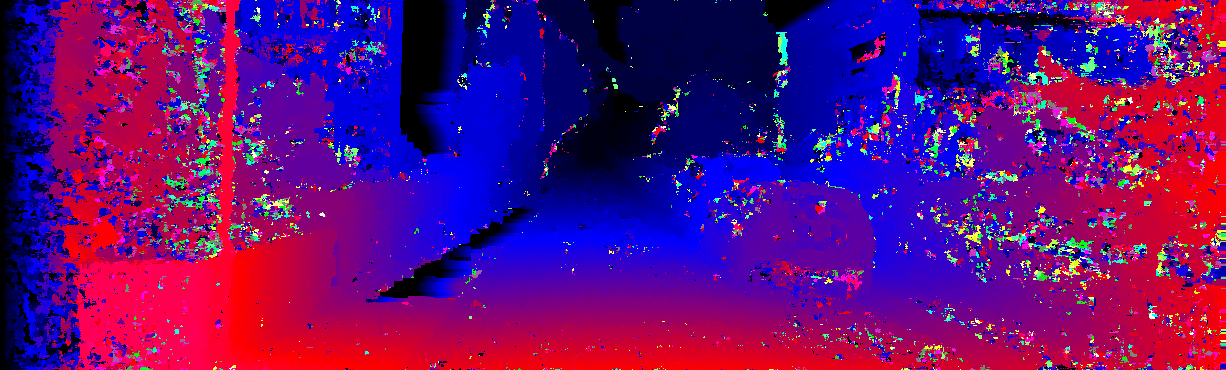} &
        \includegraphics[width=0.3\textwidth]{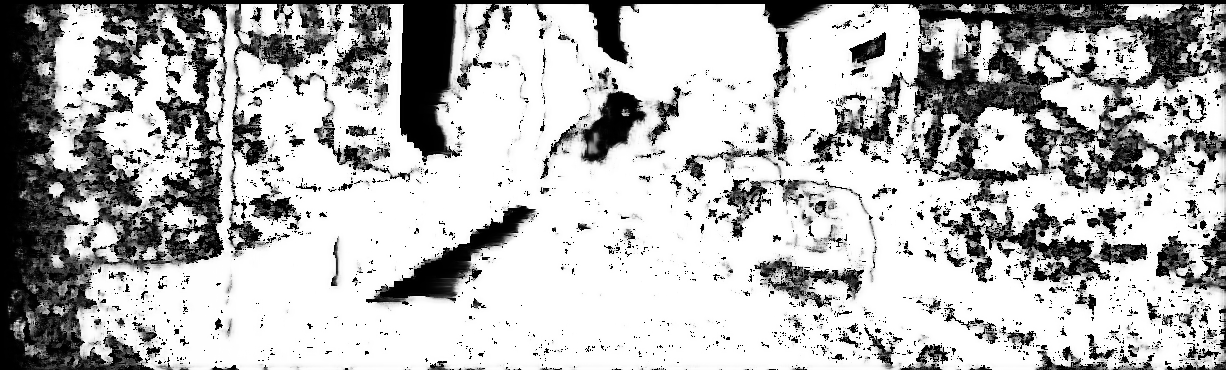} \\
    \end{tabular}
    \caption{\textbf{Example of confidence estimation.} From left to right, reference image from KITTI 2012 dataset, disparity map by MC-CNN-fst \cite{zbontar2016stereo} raw algorithm and confidence estimation inferred by LGC-Net \cite{tosi2018beyond}.}
    \label{fig:confidence}
\end{figure*}

\section{Confidence estimation}
\label{sec:confidence}

\matteo{Almost in parallel to the first attempts of replacing single steps in the stereo pipeline, learning-based confidence estimation 
\cite{poggi2017quantitative}, aiming to predict reliability of the disparity assigned to each pixel, gained popularity.} 
The first approaches relied on random forest classifiers \cite{haeusler2013ensemble,spyropoulos2014learning,spyropoulos2016correctness,park2015leveraging,park2018learning,poggi2016learning,poggi2020learning} fed with conventional (\ie not learning-based) features \cite{hu2012quantitative}, while the most recent ones on CNNs \cite{seki2016patch,poggi2016bmvc,poggi2017learning}.
Confidence estimators are typically trained on the output of a stereo algorithm using a two-class (inlier, outlier) label for each pixel, obtained from ground truth depth data by setting a threshold to distinguish between inliers and outliers in the output of the considered stereo algorithm. 
Moreover, techniques for self-supervised training of confidence estimators from video sequences \cite{Mostegel_2016_CVPR} or traditional confidence measures \cite{tosi2017learning} have been proposed in the literature.

Starting from the recent review and evaluation reported in \cite{poggi2017quantitative}, we introduce and classify novel approaches which have appeared since then. Specifically, we divide these techniques into two categories: those operating in the disparity/image domain and those processing the cost volume. \autoref{fig:confidence} shows an example of a disparity map and the associated confidence map.

\subsection{Confidence estimation in disparity/image domain}

This family of confidence estimators directly learn the reliability of each pixel from the disparity map, and optionally the reference image. \matteo{Although processing limited cues, these measures are particularly appealing when the full cost volume is not available, \eg when using off-the-shelf stereo or end-to-end models not having a cost volume at all.}

\textbf{LFN \cite{fu2018learning}.} 
Fu \etal propose to extend patch-based strategies for confidence estimation from the disparity map \cite{seki2016patch,poggi2016bmvc,poggi2017learning} in a \text{Late-Fusion Network} (LFN), which combines features extracted from both the image and disparity map, and introduce dilated convolutions to further increase the local context and potentially give more cues to the network for estimating confidence.

\textbf{LGC-Net \cite{tosi2018beyond}.} Tosi \etal extend the previous approach by considering both local cues (encoded by patches) and the global context by designing a \textit{Local Global Confidence Network} (LGC-Net), combining the large receptive field of a global sub-network with the accuracy on high-frequency noise enabled by patch-based strategies.

\subsection{Confidence estimation using the cost volume}

\matteo{Although the cost volume is often hidden from the end-user, it provides 
additional cues that are meaningful to distinguish reliable disparities from unreliable ones. For instance, it can encode the presence of multiple cost hypotheses competing for the minimum, which is information that 
cannot be retrieved from the disparity map alone.}

\textbf{Reflective confidence \cite{shaked2017improved}.}
Following the trend of replacing single steps of the stereo pipeline, Shaked and Wolf propose to jointly estimate a confidence measure together with cost optimization, before disparity selection. A two-layer fully connected network processes the matching costs, predicting confidence together with the final disparity map.

\textbf{Feature Augmentation \cite{kim2017feature}.}
As CNN based measures showed great results processing local cues from the disparity map, Kim \etal apply the same principle to confidence estimation based on random forests, by extracting a robust set of features extracted from super-pixel and concatenated with per-pixel features, similarly to \cite{gouveia15}.

\textbf{Unified Network \cite{kim2018unified}.}
Jointly learning cost optimization and confidence estimation by working on small patches proved to be effective at improving the accuracy of the final disparity map of a stereo pipeline. 
Thus, Kim \etal propose a unified architecture for cost volume optimization and confidence estimation. A first encoder-decoder module refines the matching costs with a large receptive field in order to obtain a more accurate disparity map. Then, a final sub-network processes it together with top-k refined costs to output a confidence map.

\textbf{LAF-Net \cite{Kim_2019_CVPR}.}
A larger receptive field is usually effective in improving confidence estimation \cite{kim2018unified,tosi2018beyond}. The size of the receptive field of a neural network is traditionally determined by its architecture, \ie the number of downsampling operations performed. Kim \etal develop a novel model that extracts features from the image, disparity map and cost volume to infer confidence scores. A key element of this architecture is the scale inference network, which learns the scale map and warps the fused confidence features through convolutional spatial transformer networks \cite{jaderberg2015spatial}.

\subsection{Experimental comparison}

In this section, we report a quantitative comparison between confidence estimation frameworks. We refer to experiments reported in \cite{Kim_2019_CVPR}, since it is the most recent publication that includes a fair comparison of relevant methods. All methods have been re-trained by the authors \cite{Kim_2019_CVPR} on MPI Sintel \cite{butler2012sintel} and KITTI 2012 \cite{Geiger2012CVPR}.
The Area Under the Curve (AUC) metric \cite{hu2012quantitative,poggi2017quantitative} over sparsification curves is used to evaluate the effectiveness of confidence measures with the error threshold set to 1.

\autoref{tab:confidence} reports results on KITTI 2015 and the half-resolution images of Middlebury 2014. \matteo{We also report the optimal AUC, highlighted in yellow.}
Confidence estimation is carried out on disparity maps generated by the regular Semi-Global Matching (SGM) algorithm \cite{hirschmuller08} with a census-based matching cost function and the MC-CNN-acrt pipeline \cite{zbontar2016stereo}, including SGM optimization and filtering.
The first three rows report results of top 3 methods evaluated in \cite{poggi2017quantitative}, namely O1, PBCP, CCNN. The table shows how increasing the size of the receptive fields improves confidence estimation. Indeed LGC-Net and LAF-Net are the top-performing methods on both datasets. Moreover, considering cost volume information together with image and disparity cues leads to minor improvements, enabling LAF-Net to achieve the best overall results. Nonetheless, it is worth pointing out that less constrained methods based only on image/disparity cues, such as LGC-Net, achieve very competitive results. 

This evaluation, together with those reported in previous works \cite{hu2012quantitative,poggi2017quantitative}, highlights how confidence measures are very close to optimal performance when dealing with conventional stereo algorithms, even if they include learning-based modules. However, the literature lacks papers studying confidence estimation in the case of end-to-end stereo networks. Although there are published approaches \cite{kendall2017uncertainties,kendall2018multi,Ilg_2018_ECCV,gast2018lightweight,Poggi_2020_CVPR} for estimating the uncertainty of CNNs, they have not been applied in this specific field so far, making it an exciting future research direction.

\begin{table}[t]
    \centering
    \scalebox{0.8}{
    \centering\renewcommand{\tabcolsep}{3pt}
    \begin{tabular}{|l|c|c|c|c|}
    \hline
    & \multicolumn{2}{c|}{\cellcolor{blue!5} \textbf{KITTI 2015}} & \multicolumn{2}{c|}{\cellcolor{blue!5} \textbf{Middlebury 2014}} \\
    \cline{2-5}
    \textbf{Method} & \textbf{SGM} & \textbf{MC-CNN-fst} & \textbf{SGM} & \textbf{MC-CNN-fst} \\
    \hline
    O1 \cite{poggi2016learning} & 4.61 & 2.63 & 7.91 & 7.07 \\
    PBCP \cite{seki2016patch} & 4.39 & 2.72 & 7.91 & 7.18 \\
    CCNN \cite{poggi2016bmvc} & 4.19 & 2.58 & 7.69 & 7.16 \\
    LFN \cite{fu2018learning} & 4.05 & 2.53 & 7.52 & 6.92 \\
    LGC-Net \cite{Tosi_2019_CVPR} & 3.92 & 2.36 & 7.35 & 6.85 \\
    Augment. \cite{kim2017feature} & 4.30 & 2.94 & 7.72 & 7.01 \\
    Unified \cite{kim2018unified} & 4.07 & 2.50 & 7.49 & 6.94 \\
    Reflective \cite{shaked2017improved} & 5.31 & 2.92 & 8.06 & 7.36 \\
    LAF-Net \cite{Kim_2019_CVPR} & 3.85 & 2.25 & 7.18 & 6.83 \\
    \rowcolor{LightYellow}Optimal & 3.48 & 1.70 & 5.69 & 5.27 \\
    \hline
    \end{tabular}
    }
    \smallskip
    \caption{\textbf{Experimental comparison of confidence estimators.} AUC scores ($\times10^2$) computed on KITTI 2015 and Middlebury 2014 datasets.}
    \label{tab:confidence}
\end{table}

\begin{figure*}[t]
    \centering
    \renewcommand{\tabcolsep}{1pt}
    \begin{tabular}{ccc}
        \includegraphics[width=0.3\textwidth]{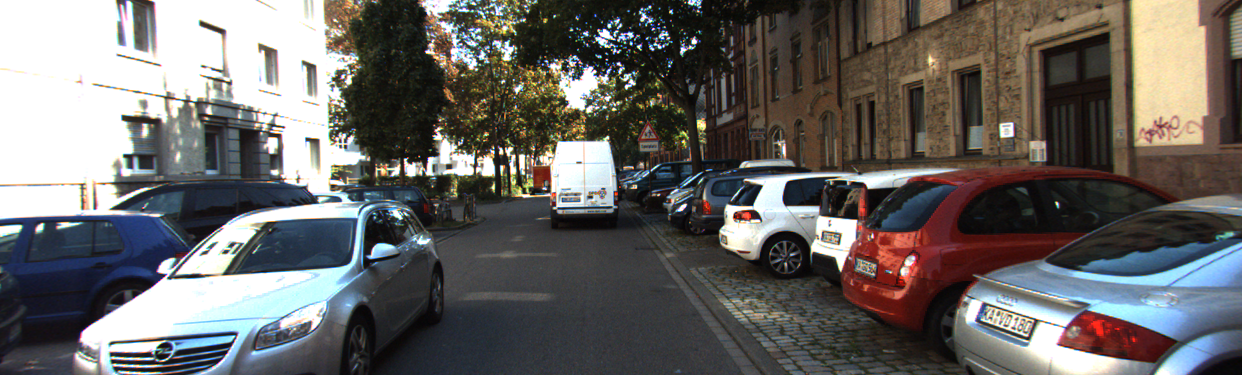} &
        \includegraphics[width=0.3\textwidth]{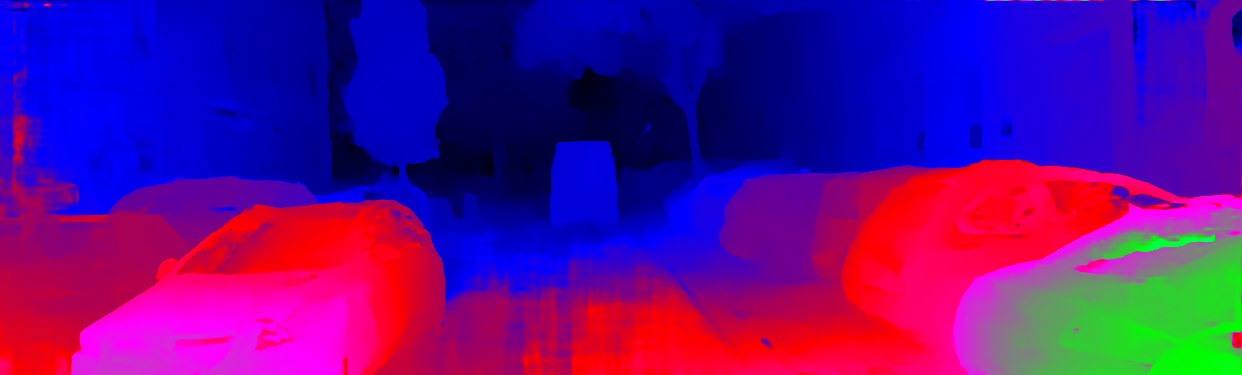} &
        \includegraphics[width=0.3\textwidth]{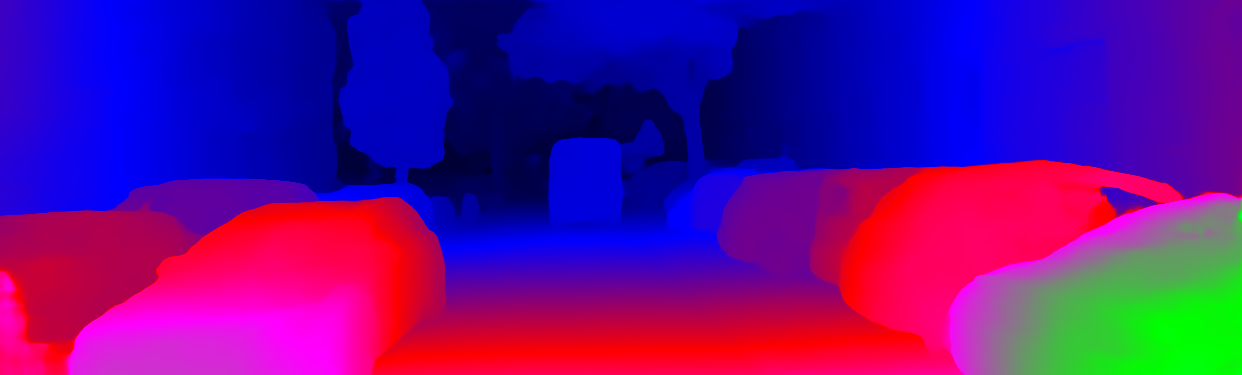} \\
    \end{tabular}
    \caption{\textbf{Effects of domain-shift.} On a KITTI 2015 stereo pair (top), a GWC-Net \cite{guo2019group} instance trained on synthetic images \cite{mayer2016large} produces poor results (middle) on the road and in reflective surfaces. A short fine-tuning on KITTI 2012 dramatically improves the results (bottom).}
    \label{fig:shift}
\end{figure*}

\section{Domain-shift: challenges and solutions}
\label{sec:shift}

We have seen that end-to-end models achieve state-of-the-art performance on most benchmarks \cite{Geiger2012CVPR,menze2015object,schops2017multi} and yield promising results on more challenging datasets \cite{scharstein2014high}, as shown in \autoref{sec:results-endtoend}.
This is made possible by training deep models on a large enough number of synthetic images to allow such complex architectures to converge. 
However, although thousands of images with different content can be easily obtained using computer graphic techniques, they currently fail at modeling many challenging properties of real imagery. In particular, the inability to accurately model camera noise, poor illumination conditions, reflections or brightness saturation produce notable loss of accuracy in disparity estimation on real stereo pairs due to the domain shift faced by the neural network.
\autoref{fig:shift} depicts an example of this phenomenon, showing the disparity maps predicted by the same model, GWC-Net \cite{guo2019group}, when trained on synthetic images only or fine-tuned on real images from the target domain.

Although theoretically possible, it is practically infeasible to collect enough images with ground truth depth for all possible real environments. To overcome this limitation, three main categories of techniques aiming to bridge the gap between synthetic and real domains have been proposed:  i) image synthesis and domain transfer, ii) self-supervised adaptation and iii) guided deep-learning.
The first category comprises methods that learn a mapping function across domains \cite{CycleGAN2017,hoffman2018cycada}, in order to make synthetic images more realistic for fine-tuning or testing. 
These techniques are general and can be applied seamlessly to different tasks; thus are beyond the scope of this survey. Therefore, we will focus on the remaining two, in \autoref{sec:adaptation} and \autoref{sec:guided}, respectively. Since different authors have deployed very different evaluation protocols, we refer to the papers for the experimental validation of each method.

\subsection{Self-supervised adaptation} 
\label{sec:adaptation}

Conversely to other tasks for which, although tedious, manual annotation is feasible in an offline manner (\eg semantic segmentation or object detection), accurately labeling images with depth measurements for training requires expensive active sensors, a cumbersome setup with appropriate calibration and often further offline refinement.

On the other hand, acquiring unlabelled image pairs is much simpler since it only requires calibrated stereo cameras. This makes the possibility of \textit{self-adapting} a neural network in the absence of ground truth labels particularly appealing, since it dramatically reduces the efforts required to bridge the gap across domains. We distinguish two main paradigms: 
\textit{offline} and \textit{online}  -- in case adaptation is carried out before encountering the new environment or directly during deployment in parallel with inference.

\subsubsection{Offline adaptation}

We classify into this category techniques that adapt a pre-trained neural network to a new domain before its deployment. These strategies are similar to conventional fine-tuning, but do not need ground truth labels.

\textbf{Confidence-guided Adaptation \cite{tonioni2017unsupervised,tonioni2019unsupervised}. } Although traditional algorithms such as SGM \cite{hirschmuller08} are widely outperformed by well-trained neural networks, they do not suffer from large drops in accuracy when applied in different domains. 
Moreover, confidence measures \cite{poggi2017quantitative} proved to be particularly effective at detecting outliers, making it possible to filter out significant errors and only keep sparse, but accurate disparity labels. 
Driven by this rationale, Tonioni \etal propose to leverage traditional (\ie not learning-based) stereo algorithms, such as SGM \cite{hirschmuller08} or \textit{block matching}, and a confidence measure, specifically CCNN \cite{poggi2016bmvc}, to process a set of unlabelled stereo pairs in order to retrieve a set of sparse, yet reliable disparity maps. Such sparse depth measurements are then used as \textit{proxy labels} for fine-tuning a deep network pre-trained on synthetic images only.

\matteo{\textbf{Self-supervised CNN-CRF \cite{knobelreiter2018self}}. Following a similar rationale, Kn\"{o}belreiter propose to improve the performance of the CNN-CRF framework \cite{Knobelreiter_2017_CVPR} to adapt to different image domains, e.g. to aerial images. In particular, they carry out a self-supervised training of the CNN in charge of computing the data term, using as labels the output of the full CNN-CRF filtered by a left-right consistency check.}

\textbf{Zoom and Learn \cite{pang2018zoom-and-learn}. }
Pang \etal observe that processing higher resolution images yields more detailed disparity maps and leverage this fact to improve the performance of models trained on different domains. In their \textit{Zoom and Learn} framework, a neural network is presented with a stereo pair and an upsampled version of the same images, thus training the model to reproduce the results achieved for the higher resolution images when lower resolution images are processed. As a result, the model learns to infer finer details from the images and thus to better tackle difficulties in generalization across domains.

\subsubsection{Online adaptation}

This paradigm removes the division between the training and testing phases. 
It \textit{continuously trains} the model any time new data are available, overcoming the need for data from the new domain before deployment.
On the other hand, adaptation starts at deployment, making early results inaccurate, but gradually improving over time.

\textbf{Open World Stereo \cite{zhong2018open}. } 
Inspired by the possibility of learning depth estimation through view synthesis \cite{godard2017unsupervised}, Zhong \etal develop \textit{Open World Stereo}, which is a 3D convolutional LSTM capable of fast convergence, typical of the 3D networks and reinforced by the memory mechanism introduced by LSTM modules. 
After an initial, \textit{prime} training phase, Open World Stereo is deployed on a new environment and, for each newly observed stereo pair, estimates an output disparity map. This latter is used at the same time to get supervision signals by warping the right image towards the left and measuring the appearance error with respect to the left image.
Although a single inference takes more than a second, the network rapidly adapts to the new environment in a few hundred iterations.

\textbf{Real-Time Self-Adaptive deep stereo \cite{Tonioni_2019_CVPR}. } The possibility of adapting \textit{on-the-fly} to a new environment is particularly appealing to make a system truly portable.
Tonioni \etal propose the first framework for real-time online self-adaptation. It relies on the synergy of two main components: i) a fast, Modularly Adaptive Network (MADNet), already introduced in \autoref{sec:2d}, and ii) an effective strategy for adapting only different portions of the entire network.
For each new incoming stereo pair, a portion is chosen according to a heuristic and supervision signals are computed through warping with the disparity estimated at the chosen resolution. Finally, back-propagation is performed only in layers belonging to the selected portion of the network. This strategy enables very fast back-propagation but, on the other hand, increases the number of steps required to converge compared to back-propagating over the entire network. The result is an approximation over time of full back-propagation that maintains real-time inference. 

\textbf{Learning to adapt \cite{Tonioni_2019_learn2adapt}. }
Efficiency, by maximizing accuracy improvement out of each adaptation step, is desirable when adapting online to new environments. 
To achieve a starting parameter configuration that is suitable for adaptation,
Tonioni \etal propose the \textit{Learning to Adapt} (L2A) training protocol. By incorporating the adaptation process itself into the learning objective through \textit{meta-learning}, the network is predisposed to a configuration of parameters that are better suited for online adaptation. This means that each offline training iteration mimics online adaptation steps over a small, synthetic sequence with ground truth. The performance gain of the adaptation phase is used as supervision for the network, leading to more efficient optimization steps.

\subsection{Guided deep learning.} 
\label{sec:guided}

This paradigm differs from previous adaptation techniques because it aims at mitigating difficulties due to domain shift without requiring explicit on-the-fly fine-tuning.
The rationale behind this approach is to \textit{guide} a network by providing external \textit{hints}. 
For instance, given sparse depth information obtained by any means (\eg LiDAR sensors) 
this method aims at overcoming the biases of a network by driving it to the correct depth values. The idea is that, since the domain shift issue is due to the appearance gap between domains (\eg synthetic vs. real images), the depth cues can overcome the loss of accuracy due to domain changes.

Poggi \etal \cite{Poggi_CVPR_2019} propose a strategy that enhances or dampens feature activations in a neural network to modulate its predictions. This is accomplished by centering a Gaussian function on each depth value provided by the sparse cues and modulating features that have a strong relationship with the output according to the Gaussian function.
These features come, respectively for 2D and 3D networks, from correlation scores or features concatenation/difference, as discussed in \autoref{sec:2d} and \autoref{sec:3d}. This technique can be applied during both training and testing to greatly reduce the effects of domain shifts.

\section{Monocular depth estimation through stereo supervision}
\label{sec:mono}

In this section, we move our focus to a new and exciting research trend: depth estimation from a single image, for which the synergy between stereo and deep learning recently allowed for results unimaginable just a few years ago. In monocular depth estimation, the goal is to learn a non-linear mapping between a single RGB image and its corresponding depth map. Even though this task comes natural to humans, it is an ill-posed problem, since a single 2D image might originate from an infinite number of different 3D scenes. However, unlike multi-view setups, a single-image depth estimation system does not require any additional equipment, making it applicable in countless scenarios. Early supervised learning approaches \cite{saxena2009make3d,ladicky2014pulling,liu2016learning,karsch2014depth,laina2016deeper} have been quite successful in this task. However, such models typically require vast amounts of pixel-wise, ground truth training annotations which are very difficult to obtain, and thus suffer from the limitations of end-to-end stereo approaches analyzed in \autoref{sec:taxonomy}.

More recent self-supervised strategies cast depth estimation as a view-synthesis problem by introducing a photometric reconstruction loss to avoid the need for expensive ground truth depth annotations. These methods received a lot of attention since they were able to take advantage of existing, or easy-to-collect, large datasets comprising either stereo pairs or monocular videos. 
\matteo{As opposed to multi-frame visual odometry, approaches based on stereo images do not suffer from scale ambiguity due to the known camera baseline.
Based on this, inferring depth without scale ambiguity results feasible even using a single RGB image as input, given the same camera at training and test time. 
This highlights once more the synergy between stereo and deep learning when dealing with depth estimation. 
In addition, with the evolution of this research trend more supervisory signals have been introduced, ranging from clever usage of the image reprojection principle to the adoption of distillation paradigms aimed at obtaining stronger loss terms. These techniques have greatly shrunk the gap with supervised approaches in terms of accuracy.}

\textbf{Geometry to the Rescue \cite{garg2016unsupervised}.} This work is the first to estimate depth from a single image using two images of a stereo pair, referred to as source $I_L$ and target $I_R$, for training.
A coarse-to-fine end-to-end convolutional neural network is trained to perform novel view synthesis, minimizing the photometric difference between the input image $I_S$ and a reconstructed one $I_W$. 
The proposed architecture infers scaled inverse depth (\ie disparity) from the source image $I_S$, which is then used to synthesize the target image $I_w$ adhering to epipolar constraints. 

\textbf{MonoDepth \cite{monodepth17}.} Other researchers follow the above seminal approach with a wide range of solutions and technical contributions. 
MonoDepth is an encoder-decoder architecture performing single image depth estimation in a self-supervised manner. Its main characteristics include i) a new training loss enforcing consistency between the predicted inverse depth maps aligned with each camera view, ii) the use of a fully sub-differentiable training loss based on the existing bilinear sampling strategy \cite{jaderberg2015spatial}, iii) a robust appearance reconstruction loss, combination of an $L1$ term and a simplified single scale SSIM \cite{SSIM} loss that compares back-warped images with their real counterparts and iv) a post-processing step to soften artifacts near occlusions, requiring two forward passes at test time. 

\textbf{AsiANet \cite{yusiong2019asianet}. } Most deep networks share the same architectural design, consisting of an encoder-decoder structure based on U-Net and various encoders, with VGG and ResNet being the most popular. 
In contrast, Yusiong \etal develop a framework dubbed \textit{Autoencoders in Autoencoders network} (AsiANet) by stacking multiple autoencoders in a multi-scale setting. Specifically, the authors employ a unique Inception-like pooling module, based on fractional max-pooling in the encoding part and multi-scale cascaded autoencoders in the decoder. This design benefits from multi-scale features when upsampling the output of the encoder taking into account local and global cues.

\textbf{3Net \cite{3net18}}. Despite the surprising effectiveness of stereo supervision, it cannot handle occlusions during training, leading monocular networks to generate artifacts in these regions. To overcome this, Poggi \etal design 3Net and supervise it by assuming three horizontally aligned images, learning to estimate two depth maps for the central frame supervised respectively by the remaining two. These two maps show occlusions in specular regions that are compensated by combining them.
Due to the lack of trinocular datasets, 3Net employs an interleaved training strategy and a custom architecture, designed to simulate a trinocular setting based on conventional binocular inputs.

\begin{table*}[!htbp]
\centering
\scalebox{0.82}{
\centering\renewcommand{\tabcolsep}{3pt}
\begin{tabular}{l|ccccc|cc|c|c|cccc|ccc}
\hline
\textbf{Method} & \textbf{S} & \textbf{V} & \textbf{P} & \textbf{A} & \textbf{GT} & \textbf{F} & \textbf{CS} & \textbf{E2E} & \textbf{Res} &  \textbf{Abs Rel} ($\bm{\downarrow}$)&  \textbf{Sq Rel} ($\bm{\downarrow}$) &  \textbf{RMSE} ($\bm{\downarrow}$) &  \textbf{RMSE log} ($\bm{\downarrow}$) &  $\bm{\delta<1.25}$ ($\bm{\uparrow}$) &  $\bm{\delta<1.25^2}$ ($\bm{\uparrow}$) & $\bm{\delta<1.25^3}$ ($\bm{\uparrow}$)\\
\hline

\hline
\multicolumn{17}{c}{\cellcolor{blue!5} \textbf{Eigen split \cite{eigen2014depth} - 697 images (maximum depth: 80m)}} \\
\hline
SfMLearner \cite{zhou2017unsupervised}  & & \checkmark & &  & & & \checkmark & \checkmark & 416 $\times$ 128 & 0.198 & 1.836 & 6.565 &  0.275 &0.718 & 0.901 & 0.960 \\
Monodepth2 \cite{godard2018digging} \textdagger & & \checkmark & & & & & & \checkmark & 1024 $\times$ 320 & 0.115 & 0.903 & 4.863 & 0.193 & 0.877 & 0.959 & 0.981 \\
\hline
Geo. to the Rescue \cite{garg2016unsupervised} & \checkmark &  & & & & &  & \checkmark & 608 $\times$ 176 & 0.152 & 1.226 & 5.849 & 0.246 & 0.784 & 0.921 & 0.967 \\
Monodepth-R50 \cite{monodepth17} & \checkmark &  &  &  & & & \checkmark  & \checkmark & 512 $\times$ 256 & 0.114 & 0.898 & 4.935 & 0.206 & 0.861 & 0.949 & 0.976 \\
AsiANet \cite{yusiong2019asianet} & \checkmark &  &  &  & & & \checkmark & \checkmark & 512 $\times$ 256 & 0.128 & 1.161 & 5.470 & 0.213 & 0.858 & 0.947 & 0.974 \\
3Net-R50 \cite{3net18} & \checkmark &  &  & &  & & \checkmark & \checkmark & 512 $\times$ 256 & 0.111 & 0.849 & 4.822 & 0.202 & 0.865 & 0.952 & 0.978 \\
MonoGAN-VGG \cite{Aleotti_monogan_2018} & \checkmark &  &   & & & & \checkmark  & \checkmark & 512 $\times$ 256 & 0.118 & 0.908 & 4.978 & 0.210 & 0.855 & 0.948 & 0.976 \\
CRF-DGAN \cite{puscas2019} & \checkmark & & & & & & & \checkmark & 512 $\times$ 256 & 0.135 & 1.182 & 5.582 & 0.235 & 0.828 & 0.933 & 0.967 \\
StrAT \cite{Mehta_3DV_2018} & \checkmark &  &  &  & & & & \checkmark & 512 $\times$ 256 & 0.128 & 1.019 & 5.403 & 0.227 & 0.827 & 0.935 & 0.971 \\
PyD-Net \cite{pydnet18} & \checkmark &  &  & & & & \checkmark  & \checkmark & 512 $\times$ 256 & 0.146 & 1.291 & 5.907 & 0.245 & 0.801 & 0.926 & 0.967 \\
SuperDepth \cite{Pillai_ICRA_2019} & \checkmark &  &  & & & & & \checkmark & 1024 $\times$ 384 & 0.112 & 0.875 & 4.958 & 0.207 & 0.852 & 0.947 & 0.977 \\
Refine and Distill \cite{Pilzer_2019_CVPR} & \checkmark & & & & & & & & 512 $\times$ 256 & 0.098 & 0.831 & 4.656 & 0.202 & 0.882 & 0.948 & 0.973 \\
Monodepth2 \cite{godard2018digging} \textdagger & \checkmark & \checkmark &  & & & & & \checkmark & 1024 $\times$ 320 & 0.106 & 0.806 & 4.630 & 0.193 & 0.876 & 0.958 & 0.980 \\
Zhan \cite{zhan2018unsupervised} & \checkmark & \checkmark &  & & & & & \checkmark & 608 $\times$ 160 & 0.135 & 1.132 & 5.585 & 0.229 & 0.820 & 0.933 & 0.971 \\
EPC++ \cite{luo2018every} & \checkmark &  &  & & & & & & 832 $\times$ 256 & 0.127 & 0.936 & 5.008 & 0.209 & 0.841 & 0.946 & 0.979 \\
\hline
DVSO \cite{yang2018deep} & \checkmark & \checkmark & \checkmark &  & & &  & & 512 $\times$ 256 & 0.097 & 0.734 & 4.442 &  0.187 &  0.888 & 0.958 & 0.980 \\
monoResMatch \cite{Tosi_2019_CVPR}  & \checkmark &  & \checkmark &  & & & \checkmark & \checkmark & 1280 $\times$ 384 & {0.096} & {0.673} & {4.351} &  {0.184} & {0.890} & {0.961} & {0.981} \\
Depth-Hints \cite{watson2018hints} \textdagger & \checkmark &  & \checkmark &  & & & & \checkmark & 1024 $\times$ 320 & 0.096 & 0.710 & 4.393 & 0.185 & 0.890 & 0.962 & 0.981\\
\hline
Kuznietsov \cite{Kuznietsov_2017_CVPR} \textdagger & \checkmark &  &  & & \checkmark & & & \checkmark & 621 $\times$ 187 & 0.113 & 0.741 & 4.621 & 0.189 & 0.862 & 0.960 & 0.986 \\
SVS \cite{luo2018single} & \checkmark &  &  & & \checkmark &  \checkmark &  & & 640 $\times$ 192 & 0.094 & 0.626 & 4.252 & 0.177 & 0.891 & 0.965 & 0.984 \\
Cross-domain \cite{guo2018learning} & \checkmark &  & \checkmark &  & \checkmark & \checkmark & & \checkmark & 512 $\times$ 256 & 0.096 & 0.641 & 4.095 & 0.168 & 0.892 & 0.967 & 0.986 \\

\hline
\multicolumn{17}{c}{\cellcolor{blue!5} \textbf{KITTI 2015 split - 40 images}} \\
\hline
Monodepth-R50 \cite{monodepth17} & \checkmark &  & & & & & \checkmark  & \checkmark & 512 $\times$ 256 & 0.159 & 2.411 & 6.822 & 0.239 & 0.830 & 0.930 & 0.967\\
SemMonodepth-R50 \cite{ramirez2018geometry} & \checkmark &  &  & \checkmark  & & & \checkmark  & \checkmark & 512 $\times$ 256 & 0.143 & 2.161 & 6.526 & 0.222 & 0.850 & 0.939  & 0.972\\
\hline
\multicolumn{17}{c}{\cellcolor{blue!5} \textbf{KITTI odometry split - 8691 images}} \\
\hline
Monodepth-R50 \cite{monodepth17} & \checkmark & & & & & & \checkmark  & \checkmark & 512 $\times$ 256 & 0.108 & 0.679 & 4.123 & 0.194 & 0.868 & 0.952 & 0.978\\
VOMonodepth-R50 \cite{vomonodepth19} & \checkmark & \checkmark & \checkmark & & & & \checkmark & \checkmark & 512 $\times$ 256 & 0.091 & 0.548 & 3.690 & 0.181 & 0.892 & 0.956 & 0.979\\
\hline 

\end{tabular}
}
\smallskip
\caption{\textbf{Quantitative evaluation on the KITTI dataset \cite{geiger2013vision}, Eigen test split \cite{eigen2014depth}.} \textdagger indicates feature extractors pre-trained on ImageNet \cite{deng2009imagenet}. 
S: Stereo, V: Video, P: Proxy, A: Additional information, GT: Ground Truth, F: Freiburg SceneFlow, CS: Cityscapes, E2E: End-to-End, Res: Resolution.
}
\label{table:eigen}
\end{table*}

\textbf{SuperDepth \cite{Pillai_ICRA_2019}}. Most of the architectures for monocular depth estimation described so far are trained using low-resolution images, due to memory constraints, and employ photometric losses. As a consequence, this forced design choice limits the attainable depth accuracy. To address this, Pillai \etal introduce a deep neural network relying on techniques typically used for super-resolution. In particular, they show that by simply increasing the input image resolution, depth estimation significantly improves accordingly. Then, taking as reference the MonoDepth \cite{godard2017unsupervised} architecture, SuperDepth obtains much better results by incorporating \textit{sub-pixel convolutional layers} to \textit{super-resolve} inverse depth maps at each scale (typically up-sampled by means of standard \textit{deconvolutions} or \textit{resize-convolution} sequences) and a \textit{differentiable flip-augmentation layer}, in order to mitigate occlusion artifacts in an end-to-end fashion.  

\textbf{SVS \cite{luo2018single}}. Although it is not a self-supervised methodology, SVS is the very first attempt to mimic stereo matching for monocular depth estimation. To this aim, processing occurs in two stages exploiting two distinct architectures.
A first view synthesis network, based on Deep3D \cite{xie2016deep3d}, is in charge of generating a synthetic right view selecting pixels from the input image $I_L$. Then, a second depth-supervised network based on the DispNet-C structure \cite{mayer2016large} processes the left and the synthesized right images to estimate the final depth map, achieving state-of-the-art results. 

\textbf{Cross-domain \cite{guo2018learning}}. Although affected by the domain-shift curse, as discussed in \autoref{sec:shift}, deep stereo networks are in general more accurate than monocular ones across domains because they reason about geometry.
Following this rationale, Guo \etal generate distilled depth labels, employing stereo networks trained on large synthetic datasets with ground truth and optionally fine-tuned on realistic ones. 
Such distilled knowledge provides dense supervision for a monocular network, resulting much more effective than warping strategies.

\textbf{MonoResMatch \cite{Tosi_2019_CVPR}}. Although distilling knowledge from stereo networks shows promise \cite{guo2018learning}, it requires a two-stage training protocol and is affected by domain-shifts. To overcome both, Tosi \etal propose respectively i) a novel architecture, jointly estimating a virtual view and performing stereo matching between it and the real image, and ii) to leverage a traditional stereo algorithm agnostic to domain, such as SGM, for distillation. As consequence, no synthetic data is required at all and a single, end-to-end training is carried out to achieve state-of-the-art accuracy.

\textbf{Refine and Distill \cite{Pilzer_2019_CVPR}}. Following distillation approaches, Pilzer \etal propose a framework for self-supervised depth estimation comprising two collaborative architectures. A \textit{student network} is in charge of synthesizing $I_W$ as opposite view to the input image (counter-intuitively, $I_R$ in this framework). Then, a backward cycle network attempts to re-synthesize the original input image taking $I_W$ as input. A \textit{teacher network} takes advantage of the inconsistencies between input image and $I_W$ to infer a refined depth map.
The last step exploits knowledge distillation, in order to transfer information from \textit{teacher} to \textit{student} and thus, to improve the \textit{student network}. 

\textbf{Depth-Hints \cite{watson2018hints}}. By studying the training signals enabled by reprojection, Watson \etal show that finding the optimal depth value is often not trivial, especially in regions of the images where the photometric loss is ambiguous and multiple depth candidates appear valid. In order to alleviate these problems, they propose to rely on external \textit{depth hints} obtained from \textit{off-the-shelf} stereo algorithms. Depth hints are trusted only when showing to be more reliable than image reprojection, thus providing complementary information during the training phase.

\textbf{MonoGAN \cite{Aleotti_monogan_2018}}. With the advent of Generative Adversarial Networks (GANs), it became possible to model distributions of complex data, enabling new capabilities in deep image synthesis for instance. Aleotti \etal use a generator network based on MonoDepth to infer depth from $I_L$ and synthesize the warped target image $I_W$. A discriminator network is then used to distinguish between \textit{fake} $I_W$ and \textit{real} $I_R$. Thus, the discriminator pushes the generator to obtain more realistic $I_W$ and, thus, better depth estimates.

\textbf{StrAT \cite{Mehta_3DV_2018}}. Following the success of GANs, this work addresses monocular depth estimation by proposing a novel \textit{Structured Adversarial Training} (StrAT) strategy. Specifically, a generator model tries to generate realistic stereo pairs, which have to be discriminated from the real ones. By incrementally varying baselines, the authors show that multiple samples with varying degrees of difficulty can be generated to guide the training process of the entire architecture.

\textbf{CRF-DGAN \cite{puscas2019}.} By studying in depth the previous GAN-based approaches, Puscas \etal propose a dual generative adversarial network, coupled with a structured Conditional Random Field (CRF), for depth prediction. In particular, two generative models infer different, complementary, disparity maps that are then fused and further processed with the outputs of two discriminative networks using the CRF. By doing so, the entire architecture establishes strong mutual constraints between each component in order to facilitate network optimization, and thus depth generation.

\textbf{PyDNet \cite{pydnet18}}. Although previous works mostly focus on accuracy, real-time and low-power constraints are often also crucial in real applications.
PyDNet represents the first attempt to make monocular depth estimation feasible on standard CPUs and embedded devices with limited memory capacity. This goal is achieved thanks to a modular design, based on a \textit{pyramidal feature extractor} and a series of shallow \textit{depth decoders}. The network infers depth maps from $\frac{1}{64}$ to half of the input resolution and allows for \textit{early-stop} at intermediate resolutions, in case of strict timing constraints. PyDNet provides results comparable to standard networks \cite{godard2017unsupervised} but considerably faster, achieving high fame rate for depth inference on embedded devices \cite{pydnet18} and on low-power platforms \cite{DATE_2019}.

\textbf{Semi-Supervised Monocular Depth \cite{Kuznietsov_2017_CVPR}.} Arguing that typical depth sensors, such as 3D laser scanners, have specific noise characteristics and generate measurements much sparser than images, while self-supervised strategies based on stereo images struggle in texture-less regions, Kuznietsov \etal propose a semi-supervised methodology to incorporate the best of both worlds. Specifically, the network exploits 3D laser measurements in supervised fashion and, at the same time, stereo pairs using a direct image alignment loss based on photometric consistency.

\textbf{Semantic MonoDepth \cite{ramirez2018}}. Semantic segmentation has witnessed enormous progress due to machine learning. 
Therefore, learning to infer both semantics and depth from a single image exploiting the synergies of the two is of particular interest. For this purpose, Ramirez \etal unite self-supervised monocular depth estimation and supervised semantic segmentation by introducing i) a shared encoder and task-specific decoders trained for joint optimization and ii) a \textit{cross-domain discontinuity loss} to enforce spatial proximity between depth discontinuities and semantic contours.  

\begin{figure*}
    \centering
    \renewcommand{\tabcolsep}{1pt}
    \begin{tabular}{cccc}
        \includegraphics[width=0.25\textwidth]{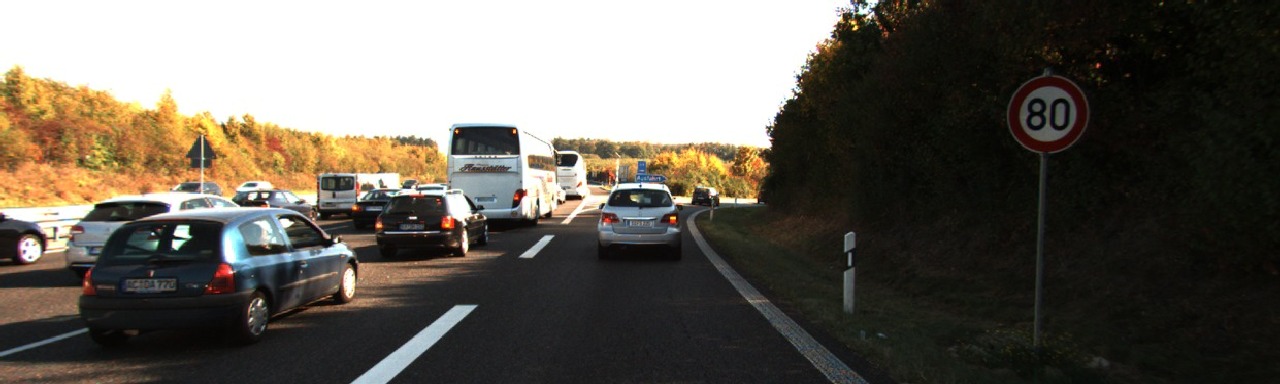} & 
        \begin{overpic}[width=0.25\textwidth]{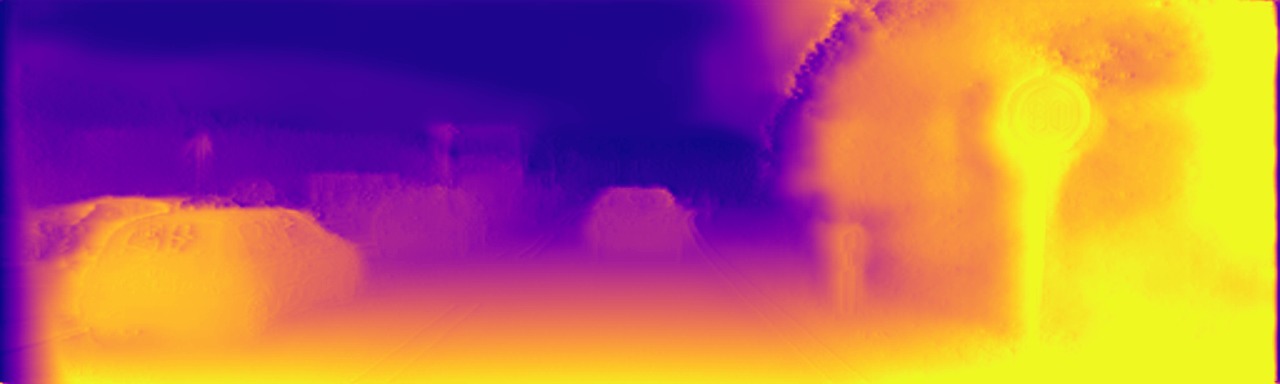}
        \put (2,22) {$\displaystyle\textcolor{white}{\textbf{2017}}$}
        \end{overpic} &
        \begin{overpic}[width=0.25\textwidth]{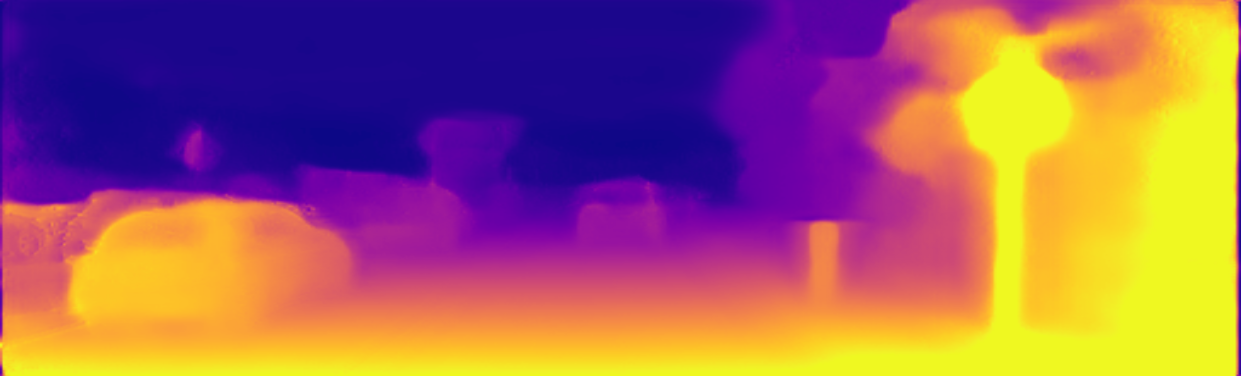}
        \put (2,22) {$\displaystyle\textcolor{white}{\textbf{2018}}$}
        \end{overpic} &
        \begin{overpic}[width=0.25\textwidth]{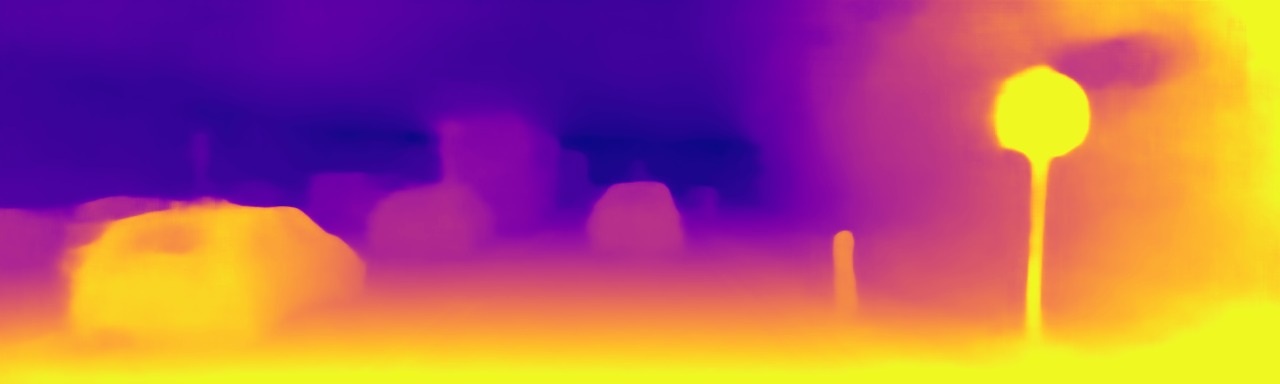}
        \put (2,22) {$\displaystyle\textcolor{white}{\textbf{2019}}$}
        \end{overpic} \\
    \end{tabular}
    \caption{\textbf{Evolution of stereo-supervised monocular depth estimation}, showing results achieved through 2017 \cite{godard2017unsupervised}, 2018 \cite{3net18} and 2019 \cite{Tosi_2019_CVPR}.}
    \label{fig:mono}
\end{figure*}

\textbf{Deep Feature Reconstruction  \cite{zhan2018unsupervised}}. An alternative path consists into learning single image depth estimation from monocular sequences, captured by a moving camera and assuming the scene to be static \cite{zhou2017unsupervised}. In this case, the pose between frames is not known, thus estimated depth suffers from scale ambiguity.
Zhan \etal propose to tackle this latter problem by using stereo sequences at training time. By imposing both spatial and temporal constraints, scene depth and camera motion are in a common real-world scale. 

\textbf{Monodepth2 \cite{godard2018digging}}. 
Most of the research on monocular depth estimation focused on complex architectures or specific loss functions. Godard \etal show that careful design choices in conjunction with light-weight models suffice to obtain state-of-the-art results. 
Considering joint supervision from a monocular sequence and a stereo pair, the authors propose i) a  minimum  reprojection  loss to effectively handle occlusions between consecutive frames, ii) a multi-scale photometric loss, upsampling low-resolution intermediate depth maps to full resolution for better supervision and iii) an auto-masking loss to ignore pixels that violate camera motion assumptions during the training phase.

\textbf{EPC++ \cite{Yang_2018_ECCV_Workshops,luo2018every}.} More authors combined supervision from sequences with stereo setup.
Luo \etal propose EPC++ (Every Pixel Counts++), taking as input two images of a monocular sequence and adopting three task-specific networks to predict camera motion, depth and optical flow. A further component, named holistic 3D motion parser (HMP), is in charge of recovering a per-pixel 3D motion for both rigid background and moving objects. 

\textbf{DVSO \cite{yang2018deep}}. To reduce the limitations of traditional, monocular visual-odometry, which is typically prone to scale drift issues due to the unknown absolute metric scale, Yang \etal incorporate dense monocular depth prediction into a monocular odometry pipeline. To this aim, the authors design a novel monocular network architecture called \textit{StackNet} and built by stacking two-subnetworks, respectively \textit{SimpleNet}, that learns to infer an initial depth estimate in a first training phase, and \textit{RefineNet}, that refines it in a second training phase. Photometric consistency losses are combined with proxy labels, sourced from a stereo odometry algorithm, in order to supervises the framework.

\textbf{VOMonodepth \cite{vomonodepth19}.} Since monocular sequences are often available during deployment too, it seems reasonable to feed a monocular network with sparse 3D cues obtained with a traditional Visual Odometry (VO) algorithm. Andraghetti \etal feed a monocular network with \textit{VO priors} to facilitate the training process and the accuracy for both complex and compact models. In contrast to strategies described so far leveraging visual odometry, it is the only solution for which external hints are available at test time as well. 

\subsection{Experimental comparison} 

\matteo{We now report a comparison 
of the monocular networks reviewed so far. To this aim, we collect performance measured on the KITTI raw dataset \cite{Geiger2012CVPR}, evaluating a set of metrics proposed by Eigen \etal{} \cite{eigen2014depth}: four error metrics, the lower the better, respectively Absolute Relative error (Abs Rel), Squared Relative error (Sq Rel), Root Mean Squared Error (RMSE) and scale-invariant RMSE (RMSE log); and three accuracy scores, obtained by counting the fraction of total pixels for which $\delta$, that is the maximum between the predicted depth to ground truth and the ground truth to depth ratios, is lower than 1.25, 1.25$^2$ and 1.25$^3$, the higher the better.}
Table \ref{table:eigen} collects results of different models on different splits of KITTI, taken from the original papers. 
For each method, we indicate the kind of supervision (S: Stereo, V: Video sequence, P: Proxy labels, A: Additional, GT: Ground Truth), additional datasets used for training (F: Freiburg SceneFlow, CS: CityScapes), the resolution adopted at training/testing time (Res) and the capability of the network to be trained in an end-to-end manner (E2E). 

Most current approaches adopt the Eigen split \cite{eigen2014depth} of KITTI, for which 697 images with ``ground truth" depth acquired with a Velodyne sensor are used for testing and 22600 for training. Methods evaluated on different splits \cite{ramirez2018geometry,vomonodepth19} are reported on bottom and compared with their baselines \cite{godard2017unsupervised}.
Depth maps are evaluated within the first 80 meters using the Garg crop \cite{garg2016unsupervised}. In this evaluation, we also consider methodologies trained only on monocular sequences \cite{zhou2017unsupervised,godard2018digging} in order to highlight the existing gap with stereo supervision and to point out how such a margin is progressively shrinking. We can observe how, in general, using proxy labels from stereo pairs or stereo sequences at training time as in \cite{yang2018deep,Tosi_2019_CVPR,Pilzer_2019_CVPR} allows to notably improve the accuracy compared to other self-supervised strategies, obtaining comparable or better results than supervised and semi-supervised networks \cite{luo2018single,guo2018learning,Kuznietsov_2017_CVPR}.
Moreover, image resolution and the pre-training process play a key role as well \cite{godard2018digging,Tosi_2019_CVPR,Pillai_ICRA_2019}.
Finally, both semantic \cite{ramirez2018geometry} and VO \cite{vomonodepth19} priors are indeed beneficial to monocular depth estimation.

Figure \ref{fig:mono} highlights the notable progress in monocular depth estimation, deploying stereo supervision, achieved in the past three years.  

\section{Discussion}
\label{sec:discussion}

\matteo{
In this section, we summarize the achievements of the recent methods reviewed in this paper and also identify the remaining open challenges and possible future research directions. 
We identify four main take-home messages:

\begin{itemize}
    \item While seminal learning-based approaches aimed at replacing single steps of the stereo pipeline showed great potential, the greatest turning-point was due to the change from hand-crafted pipelines to end-to-end networks. 
    This paradigm is nowadays dominant and represents the preferred choice for both experts and new researchers, whereas the popularity of hand-crafted pipelines is rapidly fading.
    
    \item Nevertheless, conventional knowledge about stereo survived this paradigm shift and has not gone extinct. Indeed, specific design choices such as the correlation layer \cite{mayer2016large} or 3D cost aggregation \cite{kendall2017-gcnet} are inspired by decades of research on stereo and play a key role in many deep networks.
 
    \item The main shortcomings introduced by end-to-end models concern the need for large amounts of ground truth annotated samples, that limits their seamless deployment in-the-wild. Self-supervised or adaptation techniques (\autoref{sec:adaptation}) are emerging as promising strategies to address the problem, paving the way for brand new research opportunities.
    
    \item 
    Stereo geometry turned out to also be a precious source of self-supervision for frameworks estimating depth from a single image \cite{garg2016unsupervised,godard2017unsupervised}, by dramatically reducing the overhead required to collect training samples and rapidly contributing to the spread and development of this research thrust.
    
\end{itemize}

Nevertheless, two major challenges remain in this field: i) generalization across different domains and ii) applicability on high-resolution images. In particular, current results on Middlebury 2014 \cite{scharstein2014high} highlight these open problems.
Although a few works have started addressing the former by means of continuous adaptation \cite{Tonioni_2019_CVPR} and the latter by careful design choices \cite{yang2019hsm}, in our opinion these will be the major directions of development in the upcoming years. }

\section{Conclusion}
\label{sec:conclusions}

We have presented a comprehensive survey of recent advances in 
\matteo{depth estimation from images leveraging the synergies between binocular stereo and data-driven, learning-based methods.}
Reviewing the literature reveals that the relationship is bidirectional and new machine learning approaches had to be developed to address depth estimation. While research is ongoing and voluminous, we believe that this survey will be valuable for researchers entering this field, as well as for experts.

\bibliographystyle{IEEEtran}
\bibliography{ref}

\end{document}